\documentclass[journal,twoside]{IEEEtran}

\usepackage{cite}
\usepackage{amsmath,amssymb,amsfonts}
\usepackage{algorithmic}
\usepackage{graphicx}
\usepackage{textcomp}
\usepackage[utf8]{inputenc} % allow utf-8 input
\usepackage[T1]{fontenc}    % use 8-bit T1 fonts
\usepackage{hyperref}       % hyperlinks
\usepackage{url}            % simple URL typesetting
\usepackage{booktabs}       % professional-quality tables
\usepackage{nicefrac}       % compact symbols for 1/2, etc.
\usepackage{microtype}      % microtypography
\usepackage{algorithm2e}
\usepackage{xcolor}
\usepackage{subfig}
\usepackage{caption}
\usepackage{siunitx}

\def\BibTeX{{\rm B\kern-.05em{\sc i\kern-.025em b}\kern-.08em
    T\kern-.1667em\lower.7ex\hbox{E}\kern-.125emX}}
\DeclareMathOperator{\tr}{tr}
\DeclareMathOperator{\rk}{rk}
\DeclareMathOperator{\GP}{\mathcal{GP}}

\DeclareMathOperator*{\argmax}{argmax}
\DeclareMathOperator*{\argmin}{argmin}
\DeclareMathOperator*{\diag}{diag}
\newcommand*\diff{\mathop{}\!\mathrm{d}}

\begin{document}

\title{Uncertainty-Aware Annotation Protocol to Evaluate Deformable Registration Algorithms}

\author{Lo\"ic~Peter,
		Daniel~C.~Alexander,
        Caroline~Magnain,    
        and~Juan~Eugenio~Iglesias% <-this % stops a space
\thanks{L. Peter, D. Alexander and J.E. Iglesias are with the Centre for Medical Image Computing, University College London, UK (e-mails: \{l.peter,d.alexander,e.iglesias\}@ucl.ac.uk).}% <-this % stops a space
\thanks{D. Alexander is with the NIHR UCLH Biomedical Research Centre.}
\thanks{C. Magnain and J.E. Iglesias are with the Martinos Center for Biomedical Imaging, Massachusetts General Hospital and Harvard Medical School (e-mails: \{cmagnain,jiglesiasgonzalez\}@mgh.harvard.edu).}
\thanks{J.E. Iglesias is with the Computer Science and Artificial Intelligence Laboratory (CSAIL), Massachusetts Institute of Technology.}
\thanks{This work was partially supported by the European Research Council (Starting Grant 677697), the EPSRC grant EP/M020533/1, and has been made possible in part by grant number 2019-198101 from Chan Zuckerberg Initative DAF, an advised fund of Silicon Valley Community Foundation.}
}% <-this % stops a space

\IEEEoverridecommandlockouts
\IEEEpubid{\makebox[\columnwidth]{978-1-5386-5541-2/18/\$31.00~\copyright2021 IEEE \hfill} \hspace{\columnsep}\makebox[\columnwidth]{ }}
\maketitle
\IEEEpubidadjcol

\begin{abstract}

Landmark correspondences are a widely used type of gold standard in image registration. However, the manual placement of corresponding points is subject to high inter-user variability in the chosen annotated locations and in the interpretation of visual ambiguities.
In this paper, we introduce a principled strategy for the construction of a gold standard in deformable registration. Our framework: (\textit{i})~iteratively suggests the most informative location to annotate next, taking into account its redundancy with previous annotations; (\textit{ii})~extends traditional pointwise annotations by accounting for the spatial uncertainty of each annotation, which can either be directly specified by the user, or aggregated from pointwise annotations from multiple experts; and (\textit{iii})~naturally provides a new strategy for the evaluation of deformable registration algorithms.
Our approach is validated on four different registration tasks. The experimental results show the efficacy of suggesting annotations according to their informativeness, and an improved capacity to assess the quality of the outputs of registration algorithms. In addition, our approach yields, from sparse annotations only, a dense visualization of the errors made by a registration method. The source code of our approach supporting both 2D and 3D data is publicly available at \url{https://github.com/LoicPeter/evaluation-deformable-registration}.

\end{abstract}


\IEEEpeerreviewmaketitle

\section{Introduction}

\IEEEPARstart{T}{he} accurate experimental validation of registration methods was recently identified as one of the most important remaining open problems in the field of image registration~\cite{viergever2016survey}. The main difficulty in evaluating deformable registration resides in the costly acquisition of a gold standard transformation between images, which directly follows from the high complexity of the underlying non-linear transformation. To avoid the tedious and impractical human annotation of a spatial transformation on the entire image domain, a variety of alternative evaluation strategies can be conducted~\cite{pluim2016truth}. 

A possible methodology to evaluate registration algorithms is to use synthetic data where a known simulated deformation was generated between the images~\cite{iglesias2018joint}. However, by nature, simulations are often not entirely truthful of the real-world application which is modeled, which affects the reliability of the evaluation. On real data, the quality of a predicted transformation can be indirectly quantified by annotations which are easier to acquire: for example, the delineation of some structures in two registered images enables a measure of segmentation performance (e.g., with Dice scores) which is linked to the accuracy of the registration~\cite{klein2009evaluation}. 
Such surrogate evaluation measures are particularly appropriate if they reflect the motivation behind the registration, such as in the case of registration-based segmentation. 
However, they do not sufficiently inform on the quality of the estimated transformation itself. The limitations of surrogate measures were demonstrated by Rohlfing~\cite{rohlfing2012image} who introduced adversarial algorithms able to maximize these surrogate criteria with highly inaccurate transformations. A complementary study~\cite{ribeiro2015metrics} concluded that, while surface-based metrics (such as Hausdorff distances) are more accurate than volume-based ones, none of these are able to explain more than half the variance of the true deformation field. 
Therefore, to inform directly on the true transformation, the creation of a gold standard via the manual annotation of landmark correspondences~\cite{castillo2009framework,strehlow2018landmark} remains one of the most reliable yet time-consuming solutions~\cite{rohlfing2012image}. 

To accelerate the landmark annotation process, several strategies have been proposed. Murphy~\textit{et al.}~\cite{murphy2011semi} introduced a semi-automatic approach in which, after providing a number of manual correspondences among a set of preliminarily extracted keypoints, the remaining correspondences are automatically inferred via block matching. Jegelka~\textit{et al.}~\cite{jegelka2014interactive} proposed an interactive method for annotating point correspondences within natural images, asking the user to confirm whether automatically suggested pairs of points are actual matches. The manual registration of two images can also be interactively done by using landmark annotations to constrain an automatic registration method, e.g. with Gaussian processes~\cite{fornefett2001radial,gerig2014spatially,luthi2011using} or via a joint minimization of costs respectively solving for landmark-based and deformable registration~\cite{sotiras2010simultaneous}.

Although valuable to propagate user annotations from a few landmarks to the rest of the image domain, all these approaches do not provide any principled suggestion on where the user should provide the annotations, beyond the preliminary extraction of (possibly numerous) salient keypoints. The decision on the spatial position of the provided landmarks within the images is either left to the user or driven by simple heuristics to avoid annotating landmarks that are too close to each other~\cite{jegelka2014interactive,murphy2011semi,strehlow2018landmark}. As a result, subsequent evaluations performed on these data are biased by the subjective decisions made by the user during the annotation. 
Beyond the evaluation of registration algorithms, the impact of subjective placements of landmarks has also been discussed in the context of sensor placement to perform rigid registration during surgery~\cite{shamir2012fiducial} and 
for the construction of statistical shape models~\cite{erdt2010smart,heimann2006optimal}. 

In addition, landmark-based gold standards fundamentally rely on the ability of a human user to visually identify matching locations in the two images to be registered. However, visual ambiguities can be frequently encountered in practice, such as within areas of uniform intensity or along edges, in which case the exact annotation of some landmark correspondences may not be feasible. These challenging correspondences are then either subject to an annotation error, or avoided by the annotator altogether, adding bias in the chosen locations and possibly ignoring large areas of the images. In both cases, the quality of the resulting gold standard is negatively impacted.

In this paper, we propose a principled framework for the construction of a gold standard relating two images to be registered. Building on a Gaussian process model of the true transformation, our approach iteratively suggests, in an active learning fashion, the most informative location to be annotated next in order to minimize the uncertainty on the true transformation. In addition to a landmark correspondence for each queried location, our framework supports the specification of an annotation uncertainty, either directly estimated by the annotator or obtained by merging annotations from multiple users. Based on this formalism, we also introduce a novel evaluation criterion to assess the quality of candidate transformations, typically obtained as outputs from several registration algorithms. Finally, our approach also yields a qualitative, dense visualization of the quality of a candidate transformation based on its statistical compatibility with the sparse set of provided annotations. We evaluate our approach on four different registration problems in both 2D and 3D. The results demonstrate, in a variety of settings, the benefit of our query strategy to inform on the true transformation, and a more accurate evaluation than obtained with classical landmark-based metrics.

The rest of this paper is organized as follows. In Section~\ref{sec:landmark_based_evaluation_of_deformable_registration}, we formalize the general task of image registration and of quantitative evaluation of estimated transformations. In light of this formalism, we present our methodology in Section~\ref{sec:methods} and experimentally evaluate our framework in Section~\ref{sec:experiments}.

\section{Landmark-Based Evaluation of Deformable Registration Algorithms}
\label{sec:landmark_based_evaluation_of_deformable_registration}

Let us consider two images $I_f$ and $I_m$ defined on a domain $\Omega \subset \mathbb{R}^d$, where usually $d \in \left\lbrace 2, 3\right\rbrace $, and related by a transformation $\phi : \mathbb{R}^d  \rightarrow \mathbb{R}^d$ such that we have, for all $\textbf{x} \in \Omega$, a geometric correspondence between $I_f(\textbf{x})$ and $I_m (\phi(\textbf{x}))$. The general aim of a registration algorithm is to obtain an estimate $\hat{\phi} : \mathbb{R}^d  \rightarrow \mathbb{R}^d$ as close as possible to the true transformation $\phi$ over a certain application-dependent subset of target locations $\mathcal{T} \subseteq \Omega$. For example, $\mathcal{T}$ could be given by a mask separating the tissue of interest from the background (e.g., a lung mask in~\cite{murphy2011semi}), the edge of an organ that needs to be registered accurately to build a shape model~\cite{luthi2017gaussian}, or simply set to $\Omega$ if point correspondences are desired over the whole domain. The evaluation of the quality of an estimate $\hat{\phi}$ ideally requires having knowledge of the true transformation $\phi$ in order to inspect the set of displacement errors
\begin{equation}
\Delta_{\mathcal{T}}(\phi,\hat{\phi})  = \left\lbrace \Vert \hat{\phi}(\textbf{x}) - \phi(\textbf{x}) \Vert, \textbf{x} \in \mathcal{T}\right\rbrace,
\label{eq:target_set_displacements}
\end{equation}
for example using box plots~\cite{murphy2011semi,strehlow2018landmark}. Alternatively, (\ref{eq:target_set_displacements}) can be summarized by computing statistics~\cite{castillo2009framework} such as the $p$-norms
\begin{equation}
\Vert \Delta_{\mathcal{T}}(\phi,\hat{\phi}) \Vert_p = 
\left( \frac{1}{|\mathcal{T}|} \sum_{\textbf{x} \in \mathcal{T}} {\Vert \hat{\phi}(\textbf{x}) - \phi(\textbf{x}) \Vert}^p\right)^\frac{1}{p},
\label{eq:p_norms}
\end{equation}
which covers as special cases the mean error ($p=1$), the root mean square error ($p=2$) and the maximum error ($p = \infty$).

Unfortunately, unless the set of target locations $\mathcal{T}$ is small, knowing the true value $\phi(\textbf{x})$ at every $\textbf{x} \in \mathcal{T}$ is impractical in deformable registration due to the high annotation time which would be required to manually provide spatial correspondences at every location. Therefore, the required set of displacement errors~(\ref{eq:target_set_displacements}) is not available in practice and must be approximated. Landmark-based evaluation is based on the annotation of a subset $\mathcal{L} = \left\lbrace (\textbf{x}_l , \textbf{y}_l)\right\rbrace_{1 \leq l \leq L} \subseteq \Omega^2$ of $L$ landmark correspondences at which the true transformation $\phi$ is revealed, i.e. such that $\textbf{y}_l = \phi(\textbf{x}_l)$ for all $l$. The evaluation is then conducted using
\begin{equation}
\Delta_{\mathcal{L}}(\phi,\hat{\phi})  = \lbrace \Vert \hat{\phi}(\textbf{x}_l) - \textbf{y}_l \Vert, 1 \leq l \leq L \rbrace
\label{eq:set_landmark_displacements}
\end{equation}
as a surrogate set, which appears as an approximation of~(\ref{eq:target_set_displacements}) where the target set $\mathcal{T}$ has been replaced by the set of annotated locations $\mathcal{X} =  \lbrace\textbf{x}_1, \ldots, \textbf{x}_L \rbrace$.

As a result, this approximation naturally raises the question of how to choose a set of locations $\mathcal{X}$ which best represents  $\mathcal{T}$. Moreover, using~(\ref{eq:set_landmark_displacements}) as a proxy for the true set of errors assumes that the provided annotations are true measurements satisfying $\textbf{y}_l = \phi(\textbf{x}_l)$ for all $l$. In practice, annotating a location $\textbf{x}_l$ actually results in a noisy correspondence $\textbf{y}_l \approx \phi(\textbf{x}_l)$, where the introduced error depends on the difficulty of visually assessing the point of $I_m$ which corresponds to the point $\textbf{x}_l$ in $I_f$. 
The objective of this paper is to propose a systematic methodology addressing these two sources of inter-user variability  (landmark selection and placement accuracy) in the manual construction of a gold standard for deformable registration.

\section{Methods}
\label{sec:methods}

In this section, we describe our methodology for the iterative annotation of correspondences in two images to be registered and the evaluation of transformations based on these annotations. In Section~\ref{sec:annotation_model}, we introduce a new type of annotations which extends standard pointwise correspondences. In Section~\ref{sec:transformation_prior}, we describe our probabilistic model $P(\phi)$ of the true transformation $\phi$ and how it can be conditioned on a set of manual annotations. In Section~\ref{sec:landmark_annotation}, we introduce an entropy-based criterion to suggest, at each iteration, the most informative location to annotate next. In Section~\ref{sec:efficient_implementation}, we derive mathematical results allowing an efficient implementation of this strategy. In Section~\ref{sec:uncertainty_aware_mean_square_error}, we introduce a new evaluation method based on this type of annotations and on our probabilistic model, as well as a strategy for the visualization of errors made by a registration algorithm. Finally, Section~\ref{sec:kernel_choice_and_parameter_estimation} discusses practical aspects related to the definition of suitable parameters of the transformation model for a given application.

\subsection{Annotation Model}
\label{sec:annotation_model}

We start by describing the type of annotations that we propose to collect from the user. Given a queried location $\textbf{x}_l \in \Omega$ in $I_f$, a user annotation should ideally reveal the corresponding true location $\phi(\textbf{x}_l)$ in the domain of the moving image $I_m$. Since estimating this corresponding location with certainty can often be difficult, we ask the user to provide an expected location $\textbf{y}_l$, which we assume to be a noisy version of the true value $\phi(\textbf{x}_l)$ under a $d$-variate Gaussian noise. In other words, $\textbf{y}_l$ is a realization of the random variable $\tilde{\phi}(\textbf{x}_l) = \phi(\textbf{x}_l) + \epsilon_l$ with $ \epsilon_l \sim \mathcal{N}\left(\textbf{0}, \Sigma_l\right) $ normally distributed. In addition to $\textbf{y}_l$, we also require the confidence on the annotation specified as the covariance matrix $\Sigma_l$, such that each annotation is defined as a triplet $\left( \textbf{x}_l,\textbf{y}_l,\Sigma_l\right) $.  We consider two possible approaches to obtain this covariance matrix:

\subsubsection{Graphic User Specification as an Ellipse}
The user can directly set a suitable covariance matrix from the image content by exploiting the fact that contour lines of the normal distribution are ellipses~\cite{hardle2007applied}. To do so, we query from the user a $d$-dimensional elliptic area $\mathcal{E}_l$ around the annotated central location $\textbf{y}_l$ as follows. Given a significance level $\alpha$, the user specifies a contour surface $\mathcal{E}_l$ such that the true value $\phi(\textbf{x}_l)$ is located in $\mathcal{E}_l$ with probability $1 - \alpha$. The value of $\alpha$ can be either set in advance (e.g., to $0.01$) or tailored to each annotation by the user if needed. Given a contour ellipse $\mathcal{E}_l$, we denote $\textbf{v}_1, \ldots, \textbf{v}_d$ the orthonormal basis defined by the orientation of the $d$ ellipse axes, and $r_1, \ldots, r_d \in \mathbb{R}^+$ the half-length of each axis. Then,
the covariance matrix $\Sigma_l$ corresponding to  $\mathcal{E}_l$ is
\begin{equation}
\Sigma_l =  V \diag\left[ \left( \frac{r_1}{\gamma}\right) ^2, \ldots,  \left( \frac{r_d}{\gamma}\right) ^2\right] V^\mathrm{T},
\label{eq:covariance_matrix_from_ellipse}
\end{equation}
where $V$ is the $d \times d$ matrix obtained by concatenating the $d$ vectors $\textbf{v}_1, \ldots, \textbf{v}_d$ of size $d \times 1$. The constant $\gamma$ is the solution of the equation $F(\gamma^2,d) = 1 - \alpha$, where $x \mapsto F(x,d)$ is the cumulative distribution function of the chi-squared distribution with $d$ degrees of freedom. For example, for 2D images ($d = 2$) and a $99 \%$ confidence that $\phi(\textbf{x}_l)$ is in the provided ellipse ($\alpha = 0.01$), we have $\gamma^2 \approx 9.21$. 

\begin{figure}[!t]
\centering
\subfloat[]{\includegraphics[width=0.11\textwidth]{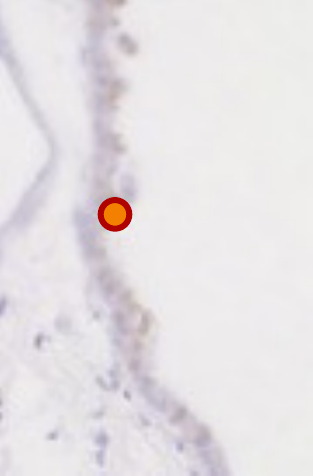}}\hfill
\subfloat[] {\includegraphics[width=0.11\textwidth]{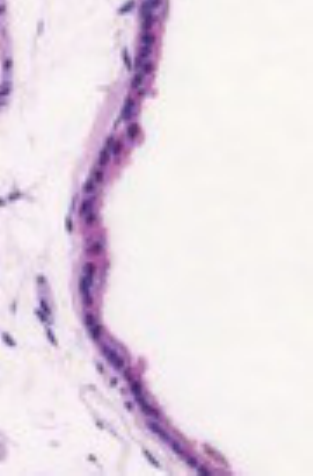}}\hfill
\subfloat[]{\includegraphics[width=0.11\textwidth]{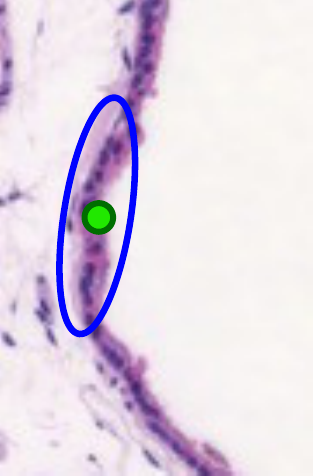}}\hfill
\subfloat[]{\includegraphics[width=0.11\textwidth]{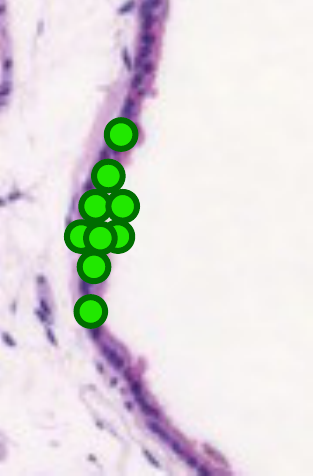}}
\caption{\textbf{Uncertain annotations.} (a)~Queried location $\textbf{x}$ within the fixed image. (b)~Corresponding area in the moving image. Although the matching point $\phi(\textbf{x})$ clearly lies somewhere along the edge, its exact location is difficult to identify visually. We present two options to obtain this anisotropic uncertainty as a bivariate Gaussian model. (c)~First option: direct manual annotation of a mean location and an elliptic confidence region around it. (d)~Second option: fusion of multiple pointwise annotations via statistical inference on the input points.}
\label{fig:landmark_ellipse_edges}
\end{figure}

With the specification of an elliptic region, the degree of confidence on each annotation can be made anisotropic, which can be especially useful for points located on edges~(Fig.~\ref{fig:landmark_ellipse_edges}). With a suitable user interface (implemented as click-and-drag interactions using OpenCV~\cite{opencv_library}), specifying a 2D ellipse only adds a small annotation overhead compared to a single click.

\subsubsection{Aggregation of Multiple Pointwise Annotations}
\label{sec:multiple_pointwise_annotations}

For an image dimensionality $d>2$, the specification of an anisotropic ellipse is not straightforward. Unless we constrain ellipses to be oriented along the coordinate axes, which would be limiting because oblique edges could then not be followed, a more advanced interface for visualization and annotation is required to place an ellipse of arbitrary directions. Therefore, we present an alternative which consists in aggregating multiple pointwise annotations from multiple raters. Retroactively, the mean location $\textbf{y}_l$ and the covariance matrix $\Sigma_l$ are obtained via a $d$-variate Gaussian fit on all annotations of the point $\textbf{x}_l$.

\subsection{Modeling Transformations with Gaussian Processes}
\label{sec:transformation_prior}

The task of deformable registration generally requires a transformation model encoding the available prior knowledge $P(\phi)$ on the true transformation or, equivalently, a regularization cost $R\left(\phi\right) $ typically linked as $P(\phi)\propto \exp\left( - R\left(\phi\right)  \right) $. Intuitively, this model defines the expected degree of smoothness of the deformation $\phi$ for a given application. To encode such prior knowledge on the transformation, Gaussian processes~\cite{rasmussen2004gaussian} have been frequently used, either as a model of the transformation itself~\cite{gerig2014spatially,jud2016sparse,luthi2017gaussian,luthi2011using,myronenko2010point,scholkopf2005object,zhu2009nonrigid}, or of the stationary velocity field parametrizing the transformation~\cite{pai2015kernel,sommer2013sparse}. The smoothness properties of $\phi$ are then defined via a suitable kernel function, possibly spatially varying~\cite{gerig2014spatially}, combining multiple kernels~\cite{sommer2013sparse}, or chosen in order to ensure theoretical guarantees such as velocity coherence~\cite{myronenko2010point}. 

We model the true non-linear transformation $\phi$ 
as a realization of a Gaussian process $\GP\left(\mu,k^{\boldsymbol{\theta}}\right)$ centered on a mean transformation $\mu : \Omega \rightarrow \mathbb{R}^d$ and where $k^{\boldsymbol{\theta}} : \Omega \times \Omega \rightarrow \bold{M}_{d \times d}(\mathbb{R})$ is a symmetric, positive-definite covariance function (also called kernel) parametrized by a vector of parameters $\boldsymbol{\theta}$. Gaussian processes generalize the multivariate normal distribution~\cite{rasmussen2004gaussian}: for any finite set of $N$ locations $\mathcal{X} = \lbrace\textbf{x}_1, \ldots, \textbf{x}_N\rbrace$, the values of $\phi$ at these locations are jointly distributed as
\begin{equation}
\phi(\mathcal{X}) \sim \mathcal{N}\left(\mu(\mathcal{X}),K^{\boldsymbol{\theta}}_{\mathcal{X} \mathcal{X}}\right),
\label{eq:gaussian_process_definition}
\end{equation}
where $\phi(\mathcal{X})$ and $\mu(\mathcal{X})$ are column vectors of size $Nd \times 1$ obtained by concatenating the $d \times 1$ vectors $\phi(\textbf{x}_1), \ldots, \phi(\textbf{x}_N)$ and $\mu(\textbf{x}_1), \ldots, \mu(\textbf{x}_N)$ respectively. The covariance matrix $K^{\boldsymbol{\theta}}_{\mathcal{X} \mathcal{X}}$ is a symmetric $Nd \times Nd$ matrix defined as
\begin{equation}
K^{\boldsymbol{\theta}}_{\mathcal{X} \mathcal{X}} =
\begin{pmatrix}
  k^{\boldsymbol{\theta}}(\textbf{x}_1,\textbf{x}_1) & \ldots & k^{\boldsymbol{\theta}}(\textbf{x}_1,\textbf{x}_N) \\
  \vdots & \vdots & \vdots\\
  k^{\boldsymbol{\theta}}(\textbf{x}_N,\textbf{x}_1) & \ldots & k^{\boldsymbol{\theta}}(\textbf{x}_N,\textbf{x}_N)
\end{pmatrix}.
\label{eq:definition_K_TT}
\end{equation}
Equation~\ref{eq:gaussian_process_definition} defines a probabilistic model $P(\phi \mid \mu, \boldsymbol{\theta})$ on the transformation, which depends on the parameters $\boldsymbol{\theta}$ and the mean transformation $\mu$. 
The choice of these parameters is application-dependent and encodes for example the acceptable amount of smoothness in the true transformation $\phi$. Here, we set $\mu$ to the identity ($\mu(\textbf{x}) = \textbf{x})$, which is a standard assumption in deformable registration corresponding to a regularization term centered on a zero displacement~\cite{gerig2014spatially}. However, if further prior knowledge is available, any other function deemed as more relevant for the given application could be chosen instead. At this stage, we assume the kernel function and its parameters to be fixed and write $k^{\boldsymbol{\theta}}$ as $k$ for clarity. The choice of our kernel and a strategy to infer $\boldsymbol{\theta}$ from a set of annotations (when needed) are given in Section~\ref{sec:kernel_choice_and_parameter_estimation}. 

Given a set of noisy annotations $\mathcal{A} = \left\lbrace (\textbf{x}_l , \textbf{y}_l, \Sigma_l) \right\rbrace_{1 \leq l \leq L}$ such that $\textbf{y}_l \sim \mathcal{N}\left(\phi(\textbf{x}_l),\Sigma_l\right)$ for all $l \in \left\lbrace 1, \ldots, L \right\rbrace $, the conditional distribution $\phi \mid \mathcal{A}$ of a Gaussian process $\phi$ is, itself, a Gaussian process $\GP\left(\mu_{\mid \mathcal{A}},k_{\mid \mathcal{A}}\right)$ with
\begin{equation}
\mu_{\mid \mathcal{A}}(\textbf{x}) = \mu(\textbf{x}) + K_{\mathcal{X}}(\textbf{x})^{\mathrm{T}} K_{\mathcal{A} \mathcal{A}}^{-1} \left( Y - \mu(\mathcal{X})\right),
\label{eq:gaussian_process_mean_conditioned}
\end{equation}
\begin{equation}
k_{\mid \mathcal{A}}(\textbf{x},\textbf{x}') = k(\textbf{x},\textbf{x}') -  K_{\mathcal{X}}(\textbf{x})^{\mathrm{T}} K_{\mathcal{A} \mathcal{A}}^{-1}  K_{\mathcal{X}}(\textbf{x}')
\label{eq:gaussian_process_covariance_conditioned}
\end{equation}
for all $\left( \textbf{x}, \textbf{x}'\right)  \in \Omega^2$~\cite{rasmussen2004gaussian}. In (\ref{eq:gaussian_process_mean_conditioned}) and (\ref{eq:gaussian_process_covariance_conditioned}), $\mathcal{X} = \lbrace\textbf{x}_1, \ldots, \textbf{x}_L\rbrace$ denotes the set of annotated locations. The $Ld \times Ld$ matrix
\begin{equation}
K_{\mathcal{A} \mathcal{A}} = K_{\mathcal{X} \mathcal{X}} + \diag(\Sigma_1, \ldots, \Sigma_L)
\end{equation} 
extends the notation $K_{\mathcal{X} \mathcal{X}}$ (defined in~(\ref{eq:definition_K_TT}) for a set of locations $\mathcal{X}$) to a set of annotations $\mathcal{A}$ by taking into account the annotation noise present in  $\mathcal{A}$. $K_{\mathcal{X}}(\textbf{x})$ is the $Ld \times d$ matrix defined as the blockwise concatenation of the $d \times d$ blocks $\left( k(\textbf{x}_l,\textbf{x})\right) _{1 \leq l \leq L}$. Finally, $Y$ and $\mu(\mathcal{X})$ are column vectors of size $Ld \times 1$ obtained by concatenating $\textbf{y}_1, \ldots, \textbf{y}_L$ and $\mu(\textbf{x}_1), \ldots, \mu(\textbf{x}_L)$ respectively.

The fact that Gaussian processes can be efficently conditioned on any set of observations $\mathcal{A}$ is crucial in our context: it allows the seamless incorporation of the collected annotations in the transformation model as the interactive process progresses, as shown for hybrid registration~\cite{gerig2014spatially,luthi2011using}. In other words, the uncertainty of $\phi$ at every location of interest is iteratively refined based on the provided correspondences $(\textbf{x}_l,\textbf{y}_l)$ and the associated confidence on the annotation, reflected by the covariance matrix $\Sigma_l$.

\begin{figure}[!t]
\centering
\includegraphics[width=0.46\textwidth]{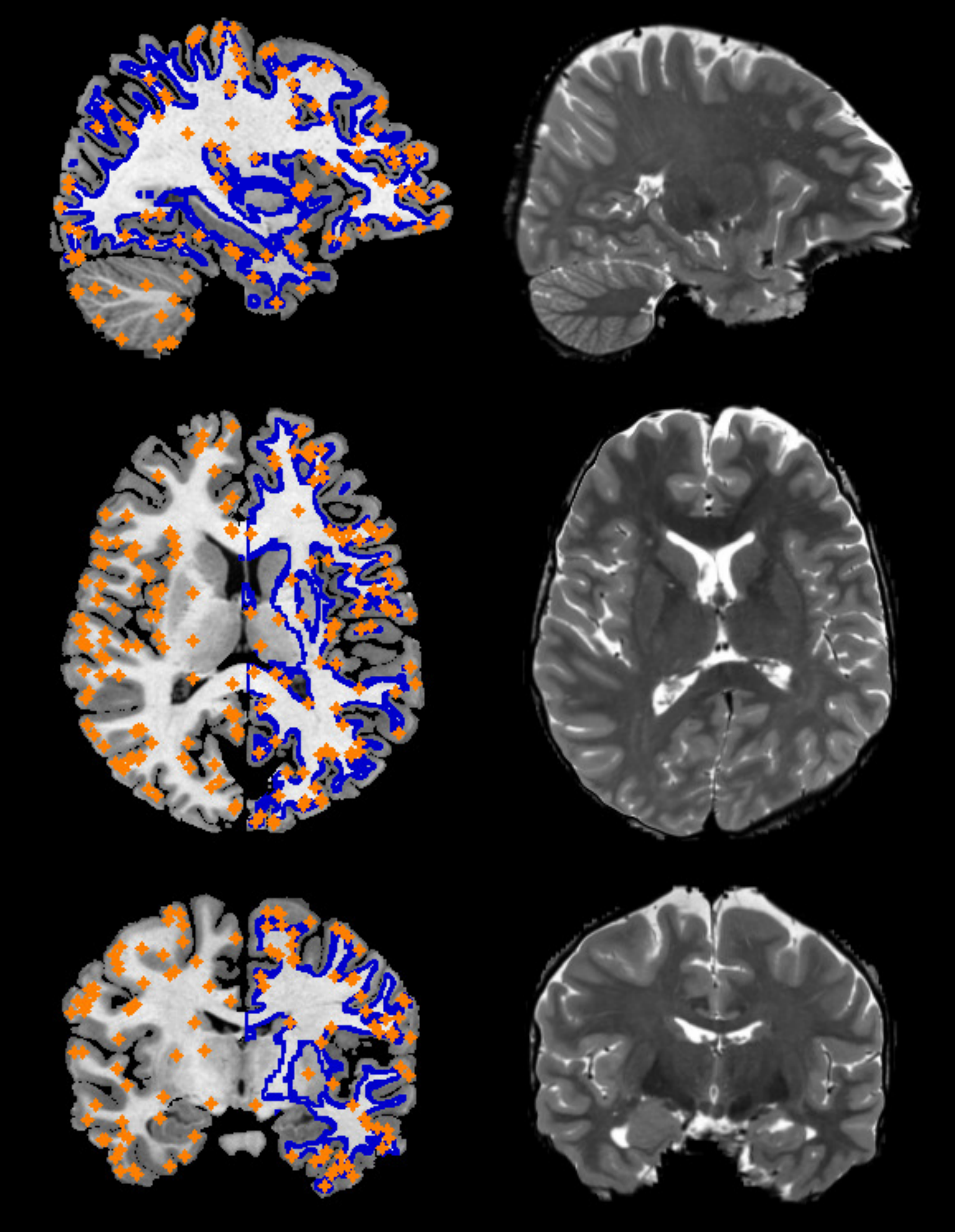}
\caption{\textbf{Example of image pair to annotate.} Left:~Three orthogonal views of an MRI volume of T1 modality serving as fixed image, together with a set $\mathcal{C}$ of extracted candidate points (shown in orange) and a target set $\mathcal{T}$ to register (cerebral white matter of the left hemisphere, shown in blue). Right:~Views of a moving image on which correspondences must be collected. This moving image was, in this case, synthetically simulated from a registered T2 image so that the true non-linear deformation relating the two volumes is densely known at every voxel for validation purposes (Section~\ref{sec:datasets}).}
\label{fig:example_t1_t2}
\end{figure}

\subsection{Suggesting the Most Informative Locations for Annotation}
\label{sec:landmark_annotation}

Our strategy for the suggestion of informative locations optionally starts with the extraction of candidate keypoints $\mathcal{C} \subseteq \Omega$ in $I_f$ (Fig.~\ref{fig:example_t1_t2}) selected for their high saliency (e.g, with a corner detector). As in the auto-completion method of Murphy \textit{et al.}~\cite{murphy2011semi}, this optional detection accelerates the annotation process by reducing the search space of the associated optimization problem (see~(\ref{eq:minimisation_problem_general}) below). 
Moreover, querying the annotation of salient locations in the fixed image makes the annotation task easier for the user, as it excludes the suggestion of locations in $I_f$ whose corresponding point in $I_m$ is bound to be difficult to find visually (such as locations in uniform areas of $I_f$). Although we do not assume any relationship between $\mathcal{C}$ and the target set $\mathcal{T}$, we show in Section~\ref{sec:efficient_implementation} that computational simplifications arise if $\mathcal{C} \subseteq \mathcal{T}$, i.e., if all candidates belong to the set of target locations.

\subsubsection{Objective Function}
To select informative locations for annotation, we seek to minimize the uncertainty on the set of displacement errors $\Delta_{\mathcal{T}}(\phi,\hat{\phi})$ between the unknown true transformation $\phi$ and an estimated transformation $\hat{\phi}$ (Section~\ref{sec:landmark_based_evaluation_of_deformable_registration}). We measure the uncertainty on the set of continuous random variables $\Delta_{\mathcal{T}}(\phi,\hat{\phi})$ with the differential entropy $\mathcal{H}$, and we seek the set of annotations $\mathcal{A}$ which minimizes $\mathcal{H}(\Delta_{\mathcal{T}}(\phi,\hat{\phi}) \mid \mathcal{A})$. Considering the transformation $\hat{\phi}$ fixed, it is mathematically equivalent to consider entropies on the set of random variables $\Phi_{\mathcal{T}} = \lbrace \phi(\textbf{x}), \textbf{x} \in \mathcal{T}\rbrace$ instead, due to invariance properties of the differential entropy~\cite{cover1991elements}. We denote $\tilde{\Phi}_{\mathcal{X}} = \lbrace \tilde{\phi}(\textbf{x}), \textbf{x} \in \mathcal{X}\rbrace$ the set of random variables provided by the user when annotating a set of locations $\mathcal{X}$, following the notations introduced in Section~\ref{sec:annotation_model}. The entropy $\mathcal{H}(\Phi_{\mathcal{T}} \mid \mathcal{A})$ corresponds to a joint entropy conditioned on this set of random variables, namely $\mathcal{H}(\Phi_{\mathcal{T}} \mid \mathcal{A}) = \mathcal{H}  (\Phi_{\mathcal{T}} \mid \tilde{\Phi}_{\mathcal{X}})$. Given a budget of $L_{\textrm{max}} \in \mathbb{N}$ correspondences that the user is ready to annotate, we ideally would like to query for annotation the set of locations
\begin{equation}
\mathcal{X} = \argmin_{\mathcal{X} \subset \mathcal{C}, |\mathcal{X}| = L_{\textrm{max}}} \mathcal{H}\left(  \Phi_{\mathcal{T}} \mid \tilde{\Phi}_{\mathcal{X}}\right).
%= \argmax_{X \subset \mathcal{C}, |X| = L_{\textrm{max}}} \mathcal{I}\left(\Phi_{\mathcal{T}},  \Phi_X\right)
\label{eq:minimisation_problem_general}
\end{equation} 

\subsubsection{Iterative Strategy}
The minimization problem defined in~(\ref{eq:minimisation_problem_general}) is a noisy version of a sensor placement problem~\cite{krause2008near} which has been shown to be NP-hard~\cite{ko1995exact}. To solve this minimization problem approximately, we follow the standard greedy iterative heuristic~\cite{krause2008near} consisting in querying, given a set of $l \geq 0$ annotated locations $\mathcal{X}_l= \left\lbrace \textbf{x}_1, \ldots, \textbf{x}_{l} \right\rbrace$, the location $\textbf{x}$ minimizing $\mathcal{H}(\Phi_{\mathcal{T}} \mid  \tilde{\Phi}_{\mathcal{X}_l \cup \left\lbrace \textbf{x} \right\rbrace })$.
Unfortunately, this is not directly feasible, since the covariance matrix governing the observation $\tilde{\phi}(\textbf{x})$ is only available after annotation of $ \textbf{x}$ by the user. To bypass this difficulty, we solve a slightly modified minimization problem and we query the location which would be the most informative \textit{if annotated perfectly}, namely
\begin{equation}
\textbf{x}_{l+1} = \argmin_{\textbf{x} \in \mathcal{C} \setminus \mathcal{X}_l} \mathcal{H}\left(  \Phi_{\mathcal{T} \setminus \left\lbrace \textbf{x} \right\rbrace } \mid  \tilde{\Phi}_{\mathcal{X}_l} , \Phi_{\left\lbrace \textbf{x} \right\rbrace}\right).
\label{eq:minimisation_problem_greedy}
\end{equation}
Note that we considered in~(\ref{eq:minimisation_problem_greedy}) the entropy on the restricted set $\Phi_{\mathcal{T} \setminus \left\lbrace \textbf{x} \right\rbrace }$ instead of $\Phi_{\mathcal{T}}$ since the differential entropy $\mathcal{H}\left( \Phi_{\mathcal{T}} \mid \tilde{\Phi}_{\mathcal{X}_l} , \Phi_{\left\lbrace \textbf{x} \right\rbrace}\right)$ diverges if $\textbf{x} \in \mathcal{T}$.

Beyond facilitating the resolution of~(\ref{eq:minimisation_problem_general}), adopting an iterative strategy offers additional advantages: (\textit{i})~instead of requiring an annotation budget $L_{\textrm{max}}$, the annotation process can be adaptively stopped, e.g., when a certain threshold on the uncertainty is reached; (\textit{ii})~the uncertainty on the annotated location, if immediately revealed by the user, can be taken into account for the following suggestions; and (\textit{iii})~the  parameters $\mu$ and $\boldsymbol{\theta}$ encoding the Gaussian process prior on $\phi$ can be revised based on the user annotations if necessary (see Section~\ref{sec:kernel_choice_and_parameter_estimation}).

\subsubsection{Entropy of a Gaussian process}
The entropy-based iterative strategy~(\ref{eq:minimisation_problem_greedy}) is generic for any transformation model. Under the Gaussian process assumption made in Section~\ref{sec:transformation_prior}, the entropies in~(\ref{eq:minimisation_problem_greedy}) can be computed as follows. For any finite sets of target locations $\mathcal{T} = \lbrace\textbf{t}_1, \ldots, \textbf{t}_N\rbrace $ and annotations $\mathcal{A}$ acquired at the locations $\mathcal{X}$, we have (see e.g.,~\cite{rasmussen2004gaussian})
\begin{equation}
\mathcal{H}\left(  \Phi_{\mathcal{T}} \mid \mathcal{A} \right) = \frac{d |\mathcal{T}|}{2} \log (2 \pi e) + \frac{1}{2} \log\left( \det K_{\mathcal{T} \mathcal{T} \mid \mathcal{A}}\right)
\label{eq:entropy_gp}
\end{equation}
where, by analogy with~(\ref{eq:gaussian_process_covariance_conditioned}), we define
\begin{equation}
K_{\mathcal{T} \mathcal{T} \mid \mathcal{A}} = K_{\mathcal{T} \mathcal{T}} - K_{\mathcal{X} \mathcal{T}}^\mathrm{T}  K_{\mathcal{A} \mathcal{A}}^{-1} K_{\mathcal{X} \mathcal{T}},
\label{eq:definition_K_TT_given_A}
\end{equation} 
where the matrix $K_{\mathcal{X} \mathcal{T}} = \left( k(\textbf{x}_i,\textbf{t}_j)\right)_{i,j}$ is of size $Ld \times Nd$.

Remarkably, as apparent from~(\ref{eq:gaussian_process_covariance_conditioned}), the conditioned covariance function $k_{\mid \mathcal{A}}$ is, unlike $\mu_{\mid \mathcal{A}}$, independent of the annotations $\textbf{y}_1, \ldots, \textbf{y}_L$ provided on the moving image by the user. As a result, the entropy~(\ref{eq:entropy_gp}) does not depend on these manual annotations either. Therefore, if an estimate of the annotation uncertainty cannot be specifed at annotation time (and thus is set to a fixed isotropic value instead), the series of suggested annotations \textit{does not depend on the user input}. For this reason, our suggestion approach can seamlessly combine the annotations of multiple annotators which can be retroactively fused to obtain uncertainty estimates on each queried location, as we proposed in Section~\ref{sec:multiple_pointwise_annotations}.

\begin{algorithm}[t]
\small	
\SetKwInOut{Input}{Inputs}
\SetKwInOut{Output}{Output}
	\Input{Images $I_f$ and $I_m$, Annotation budget $L_{\textrm{max}}$, Target set $\mathcal{T}$ within $I_f$, Candidate set $\mathcal{C}$ within $I_f$, Kernel function $k$ and its parameters;}
\textbf{Initialization:} $\mathcal{X}_0 \leftarrow \emptyset$; $\mathcal{A}_0 \leftarrow \emptyset$;\\
 Precompute $K_{\mathcal{T}\mathcal{T}}^{-1}$ (can be done offline);\\
 \For{$l \leftarrow 0 \ \KwTo \ L_{\textrm{max}}-1$} 
 {
 	\For{$\bold{x} \in \mathcal{C} \setminus \mathcal{X}_l$}
 	{
 	\eIf{$\bold{x} \in \mathcal{T}$}
 		{Compute $\Delta \mathcal{H}_l\left( \textbf{x}\right)$ with~(\ref{eq:delta_H_in_T});
 		}
 		{Compute $\Delta \mathcal{H}_l\left( \textbf{x}\right)$ with~(\ref{eq:delta_H_not_in_T});}
 	}
 	$\bold{x}_{l+1} \leftarrow \argmax_{\bold{x} \in \mathcal{C} \setminus \mathcal{X}_l} \Delta \mathcal{H}_l\left( \textbf{x}\right)$;\\
  Query annotation of $\bold{x}_{l+1}$\; 
  The user provides $(\bold{y}_{l+1}, \Sigma_{l+1})$;\\
  $\mathcal{X}_{l+1} \leftarrow \mathcal{X}_{l} \cup \left\lbrace \bold{x}_{l+1} \right\rbrace$\;
  $\mathcal{A}_{l+1} \leftarrow \mathcal{A}_{l} \cup \left\lbrace \left( \bold{x}_{l+1}, \bold{y}_{l+1}, \Sigma_{l+1} \right) \right\rbrace$;\\
Compute $K_{\mathcal{A}_{l+1} \mathcal{A}_{l+1}}^{-1}$ from $K_{\mathcal{A}_l \mathcal{A}_l}^{-1}$;\\
  Compute $K_{\mathcal{T} \cup \mathcal{A}_{l+1} \mathcal{T} \cup\mathcal{A}_{l+1}}^{-1}$ from $K_{\mathcal{T} \cup \mathcal{A}_l \mathcal{T} \cup\mathcal{A}_l}^{-1}$;

 }
 	\Output{Annotations $\mathcal{A}_{L_{\textrm{max}}}$}\
 \caption{\textbf{Pseudo-code of our proposed annotation strategy.} The matrices $K_{\mathcal{T}\mathcal{T}}^{-1}$ and $K_{\mathcal{T} \cup \mathcal{A}_l \mathcal{T} \cup\mathcal{A}_l}^{-1}$ are only used in~(\ref{eq:delta_H_not_in_T}) to compute a score on the locations which both belong to $\mathcal{C}$ and do not belong to $\mathcal{T}$. If the evaluation task is defined such that $\mathcal{C} \subseteq \mathcal{T}$, the computation of these matrices is therefore not needed and an arbitrarily large target set can then be handled without computational overhead.}
\label{fig:algorithm}
\end{algorithm}

\begin{figure*}
\centering
\subfloat[Fixed image]{\includegraphics[width=0.32\textwidth]{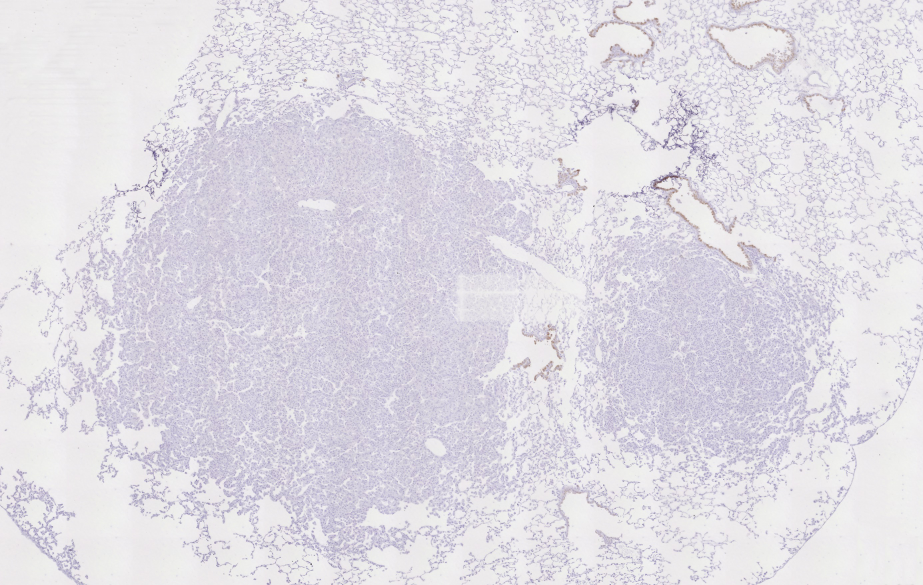}}\hfill
\subfloat[Registered moving image ($\hat{\phi}_0$)]{\includegraphics[width=0.32\textwidth]{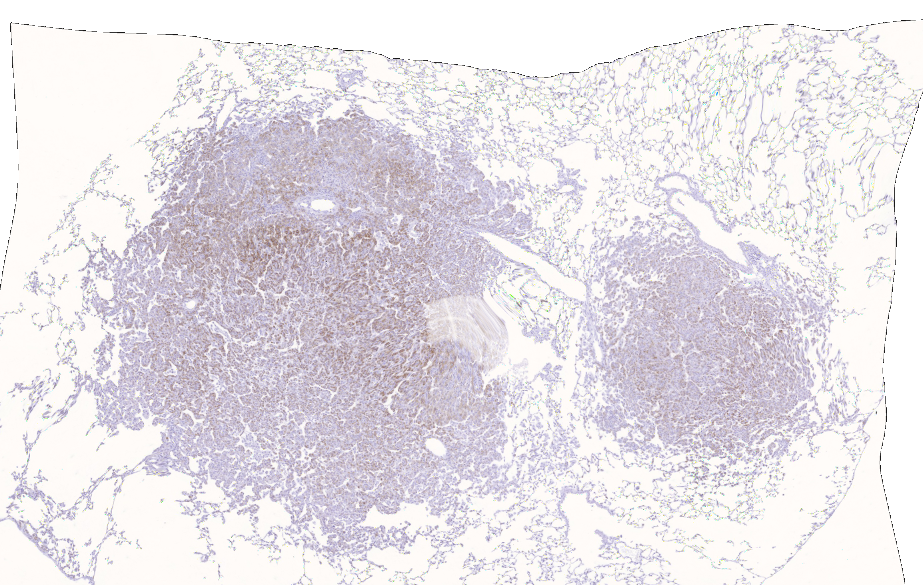}}\hfill
\subfloat[
Entropy map after $96$ annotations] {\includegraphics[width=0.32\textwidth]{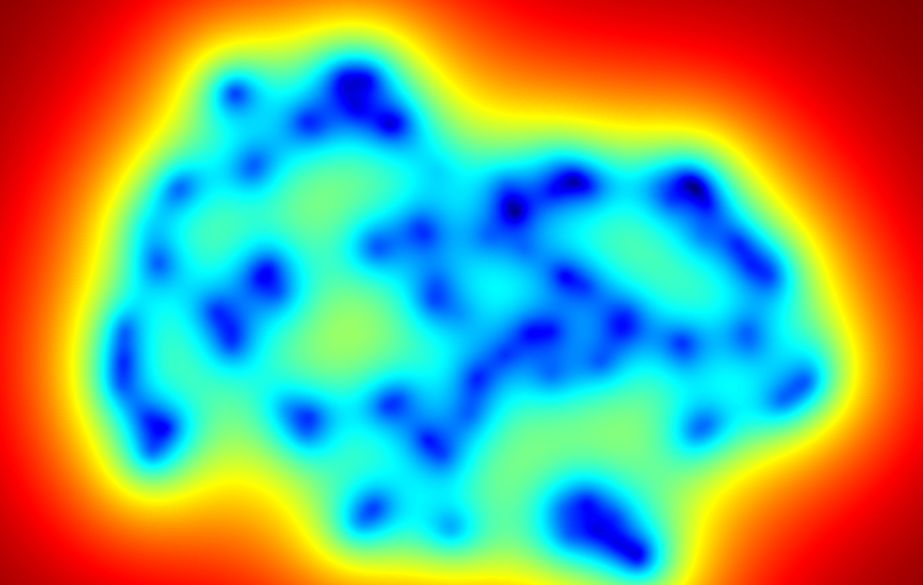}}\hfill
\subfloat[Corresponding error map of $\hat{\phi}_0$]{\includegraphics[width=0.32\textwidth]{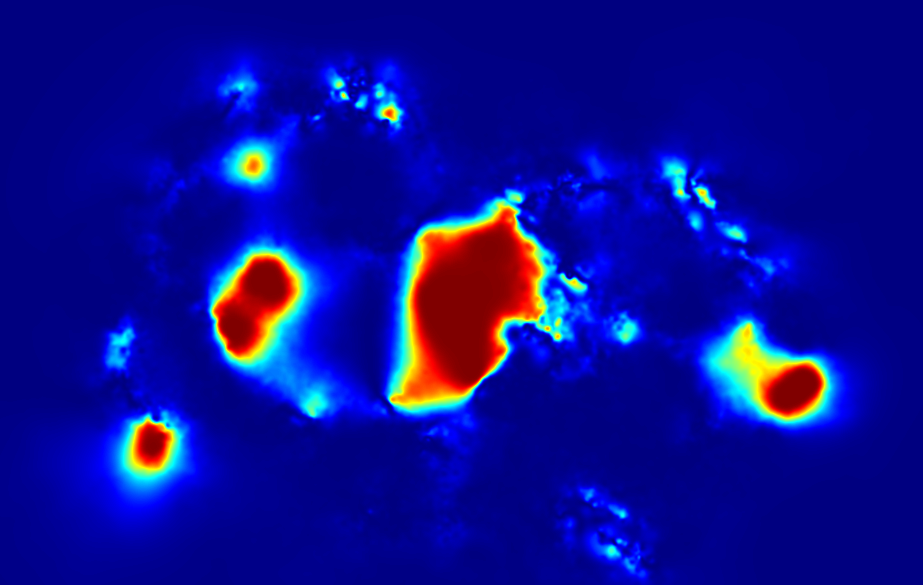}}\hfill
\subfloat[Error map of $\hat{\phi}_1$] {\includegraphics[width=0.32\textwidth]{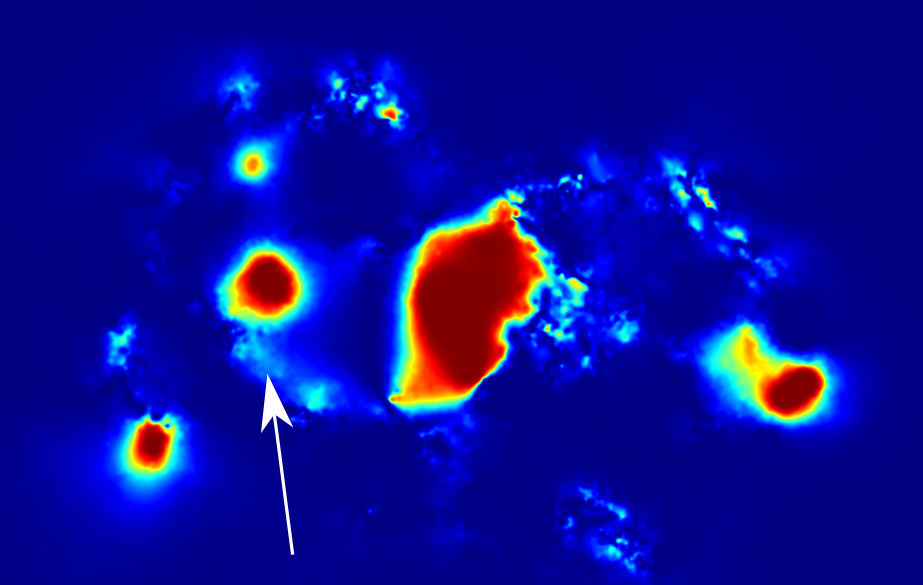}}\hfill
\subfloat[Combined visualization for $\hat{\phi}_0$\label{fig:blended_map}]{\includegraphics[width=0.32\textwidth]{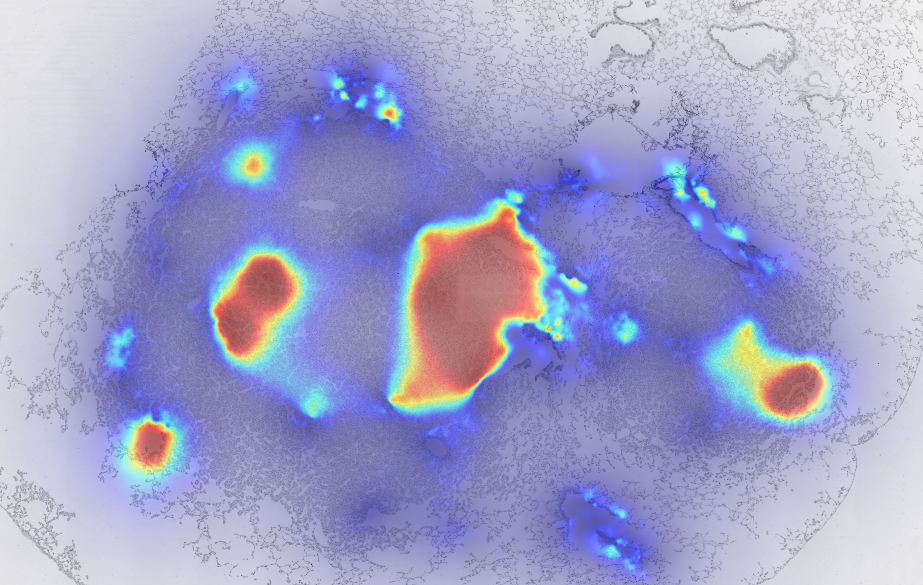}}\hfill
\caption{\textbf{Visualization of errors made by a registration algorithm.} Given a pair of annotated images for which a registration algorithm was run, our probabilistic framework allows a quick visualization of misregistered areas by displaying the areas of the domain which statistically contradict the annotations. (a)~Fixed image from the CIMA dataset. (b)~Registered moving image. We denote $\hat{\phi}_0$ the corresponding estimated transformation. (c)~Entropy map after manually reannotating, with user-specified ellipses, the 96 correspondences originally available on this pair. (d) Error map of $\hat{\phi}_0$. (e)~Error map for another estimated transformation $\hat{\phi}_1$. The fact that $\hat{\phi}_1$ is more accurate than $\hat{\phi}_0$ in a particular area (see white arrow) can be easily visualized from the two error maps. (f)~Example of blended map combining (c) and (d) by rendering high entropy areas as transparent. Among the blue areas of the error map (d), the blended visualization yields a visual distinction between the areas where the registration is indeed accurate and the areas that were not sufficiently annotated to be able to conclude on the registration accuracy.}
\label{fig:example_heat_map}
\end{figure*}

\subsection{Efficient Iterative Implementation}
\label{sec:efficient_implementation}

In order to solve~(\ref{eq:minimisation_problem_greedy}), the entropy $\mathcal{H}(  \Phi_{\mathcal{T} \setminus \left\lbrace \textbf{x} \right\rbrace } \mid  \tilde{\Phi}_{\mathcal{X}_l} , \Phi_{\left\lbrace \textbf{x} \right\rbrace})$ must be computed for each candidate  $\textbf{x} \in \mathcal{C} \setminus \mathcal{X}_l$. Evaluating these entropies using~(\ref{eq:entropy_gp}) requires two computationally expensive steps per candidate $\textbf{x}$: 
the inversion of a matrix of size $(l+1)d \times (l+1)d$ in~(\ref{eq:definition_K_TT_given_A}), and the computation of a $(N-1)d \times (N-1)d$ determinant in~(\ref{eq:entropy_gp}).
For a large number  $l$ of available annotations or for a large number $N$ of target points, these computational bottlenecks can lead to a non-negligible waiting time between suggestions, which would compromise the interactivity of the algorithm. Fortunately, an efficient implementation and real-time processing between suggestions can be obtained by noticing that, for all $\textbf{x} \in \mathcal{C} \setminus \mathcal{X}_l$,
\begin{equation}
\mathcal{H}\left(  \Phi_{\mathcal{T} \setminus \left\lbrace \textbf{x} \right\rbrace } \mid  \tilde{\Phi}_{\mathcal{X}_l} , \Phi_{\left\lbrace \textbf{x} \right\rbrace}\right) = \mathcal{H}\left(  \Phi_{\mathcal{T}} \mid  \tilde{\Phi}_{\mathcal{X}_l}\right) - \Delta \mathcal{H}_l \left( \textbf{x}\right),
\label{eq:fast_optimization}
\end{equation}
where $\Delta \mathcal{H}_l\left( \textbf{x}\right)$ is given by:
\begin{itemize}
\item If $\textbf{x} \in \mathcal{T}$,
\begin{equation}
\Delta \mathcal{H}_l\left( \textbf{x}\right)  = \mathcal{H}\left( \Phi_{\left\lbrace \textbf{x} \right\rbrace} \mid \tilde{\Phi}_{\mathcal{X}_l}  \right).
\label{eq:delta_H_in_T}
\end{equation}
\item If $\textbf{x} \notin \mathcal{T}$, 
\begin{equation}
\Delta \mathcal{H}_l\left( \textbf{x}\right)  = \mathcal{H}\left( \Phi_{\left\lbrace \textbf{x} \right\rbrace} \mid \tilde{\Phi}_{\mathcal{X}_l}  \right) - \mathcal{H}\left( \Phi_{\left\lbrace \textbf{x} \right\rbrace} \mid \Phi_{\mathcal{T}} , \tilde{\Phi}_{\mathcal{X}_l} \right).
\label{eq:delta_H_not_in_T}
\end{equation}
\end{itemize}
Equation~(\ref{eq:fast_optimization}) is a consequence of the chain rule on entropies~\cite{cover1991elements} stating that, for random variables $X$ and $Y$ (or sets of random variables), the equality
\begin{equation}
\mathcal{H}\left(X,Y\right) = \mathcal{H}\left(X \mid Y\right) + \mathcal{H}\left(Y\right)
\label{eq:conditional_entropy}
\end{equation}
holds. Since, in~(\ref{eq:fast_optimization}), $\mathcal{H}\left(  \Phi_{\mathcal{T}} \mid  \tilde{\Phi}_{\mathcal{X}_l}\right)$ does not depend on $\textbf{x}$, our query strategy~(\ref{eq:minimisation_problem_greedy}) can be rewritten as
\begin{equation}
\textbf{x}_{l+1} = \argmax_{\textbf{x} \in \mathcal{C} \setminus \mathcal{X}_l} \Delta \mathcal{H}_l\left( \textbf{x}\right),
\label{eq:maximisation_problem_greedy_simplified}
\end{equation}
which can be solved efficiently, as the determinants involved in the computation of $\Delta \mathcal{H}_l\left( \textbf{x}\right)$ are of size $d \times d$ and thus very fast to compute. Although computing the entropies in~(\ref{eq:delta_H_in_T}) and~(\ref{eq:delta_H_not_in_T}) still requires inverting two matrices $K_{\mathcal{A}_l \mathcal{A}_l}$ and $K_{\mathcal{T} \cup \mathcal{A}_l \mathcal{T} \cup \mathcal{A}_l}$, these two matrices do not depend on the location $\textbf{x}$. Therefore, their inverses must be computed only once per iteration $l$ and can, moreover, be efficiently updated from $K_{\mathcal{A}_{l-1} \mathcal{A}_{l-1}}^{-1}$ and $K_{\mathcal{T} \cup \mathcal{A}_{l-1} \mathcal{T} \cup \mathcal{A}_{l-1}}^{-1}$  by exploiting their $2 \times 2$ block structure~\cite{rasmussen2004gaussian}. 
An overview of our iterative strategy is given in Algorithm~\ref{fig:algorithm}.

As apparent on (\ref{eq:delta_H_in_T}), the score $\Delta \mathcal{H}_l\left( \textbf{x}\right)$ does not depend on $\mathcal{T}$ if $\textbf{x} \in \mathcal{T}$. Therefore, if $\mathcal{C} \subseteq \mathcal{T}$,
the queried locations are mathematically independent of the target set $\mathcal{T}$, leading to additional computational simplifications (see Algorithm~\ref{fig:algorithm}). This special case may either arise naturally from the application (e.g., if $\mathcal{T} = \Omega$ or if $\mathcal{T}$ is a foreground mask) or be artificially enforced by considering either a smaller set of candidates $\mathcal{C}' = \mathcal{C} \cap \mathcal{T}$, or an extended target set $\mathcal{T}' = \mathcal{T} \cup \mathcal{C}$.

\subsection{Uncertainty-Aware Evaluation of Registration Algorithms}
\label{sec:uncertainty_aware_mean_square_error}

\subsubsection{Uncertainty-Aware Evaluation Score}

After completion of the annotation process, a set $\mathcal{A}$ of user-provided landmark correspondences and their annotation uncertainty is available. Our Gaussian process transformation model, conditioned on the set of annotations $\mathcal{A}$, yields a probability distribution of the true transformation at each target location $\textbf{x} \in \mathcal{T}$, namely
\begin{equation}
\phi(\textbf{x}) \mid \mathcal{A} \sim \mathcal{N}\left(\mu_{\mid \mathcal{A}}(\textbf{x}),k_{\mid \mathcal{A}}(\textbf{x},\textbf{x})\right),
\label{eq:conditioned_distribution_on_true_transformation}
\end{equation}
as defined in Section~\ref{sec:transformation_prior}. Going back to the objective of landmark-based evaluation of deformable registration described in Section~\ref{sec:landmark_based_evaluation_of_deformable_registration}, we leverage our model to introduce a new approximation of the true, unavailable set $\Delta_{\mathcal{T}}(\phi,\hat{\phi})$ as:
\begin{equation}
\Delta_{\mathcal{T}}(\phi,\hat{\phi} \mid \mathcal{A})  = \left\lbrace \Vert \hat{\phi}(\textbf{x}) - \mu_{\mid \mathcal{A}}(\textbf{x}) \Vert, \textbf{x} \in \mathcal{T}\right\rbrace,
\label{eq:target_set_displacements_on_annotations}
\end{equation}
i.e., we substitute our mean Gaussian process prediction to the true transformation. Our experimental results demonstrate that an evaluation based on this set improves over the standard evaluation on the annotated locations only (see Section~\ref{sec:experiments_capacity_evaluation}).

\subsubsection{Heat Map Visualization of Registration Errors}

The probabilistic information at each location $\textbf{x}$ given by~(\ref{eq:conditioned_distribution_on_true_transformation}) defines the likelihood of any predicted value $\hat{\phi}(\textbf{x})$ estimated by a registration algorithm. Using statistical testing, we can quantify the statistical deviations of a given value $\hat{\phi}(\textbf{x})$ from the true distribution (conditioned on the annotations) given by~(\ref{eq:conditioned_distribution_on_true_transformation}). More precisely, for $\textbf{x} \in \Omega$, we consider the p-value of the $\chi^2(d)$ distribution which governs the squared Mahalanobis distance between $\hat{\phi}(\textbf{x})$ and $\mu_{\mid \mathcal{A}}(\textbf{x})$, and define the error score 
\begin{equation}
e(\textbf{x} \mid \hat{\phi}, \mathcal{A} ) = F\left( \sum_{i=1}^d  \frac{\left<\textbf{v}_i(\textbf{x}) ,\hat{\phi}(\textbf{x}) - \mu_{\mid \mathcal{A}}(\textbf{x})\right>^2}{\sigma_i^2(\textbf{x})}, d\right),
\end{equation}
where the $\sigma_i^2(\textbf{x})$ and $\textbf{v}_i(\textbf{x})$ are respectively the eigenvalues and eigenvectors of $k_{\mid \mathcal{A}}(\textbf{x},\textbf{x})$, and $x \mapsto F(x,d)$ is the cumulative distribution function of the $\chi^2(d)$ distribution. By computing the error score $e(\textbf{x} \mid \hat{\phi}, \mathcal{A} )  \in \left[0,1\right]$ at every location $\textbf{x}$ of the image domain, we obtain a dense and interpretable visualization of the errors made by a registration algorithm, defined as the statistical incompatibilities of the predicted transformation with the provided annotations (Fig.~\ref{fig:example_heat_map}).

The interpretation of these error maps is subject to the standard considerations on statistical testing. A location where the score is high always signals a deviation from the provided annotations and thus a registration error. However, a low score can be either due to an accurate registration or to insufficient data, i.e, high uncertainty at this location. For an improved interpretation of error maps, it is thus useful to consider them jointly with the entropy profile which indicates the uncertainty of the Gaussian process on the image domain (Fig.~\ref{fig:blended_map}).

\subsection{Specification of the Kernel Function}
\label{sec:kernel_choice_and_parameter_estimation}

We conclude the section with additional details on the specification of a Gaussian process in practice.

\subsubsection{Chosen Kernel Function}
\begin{figure}[t]
\centering
\includegraphics[width=0.46\textwidth]{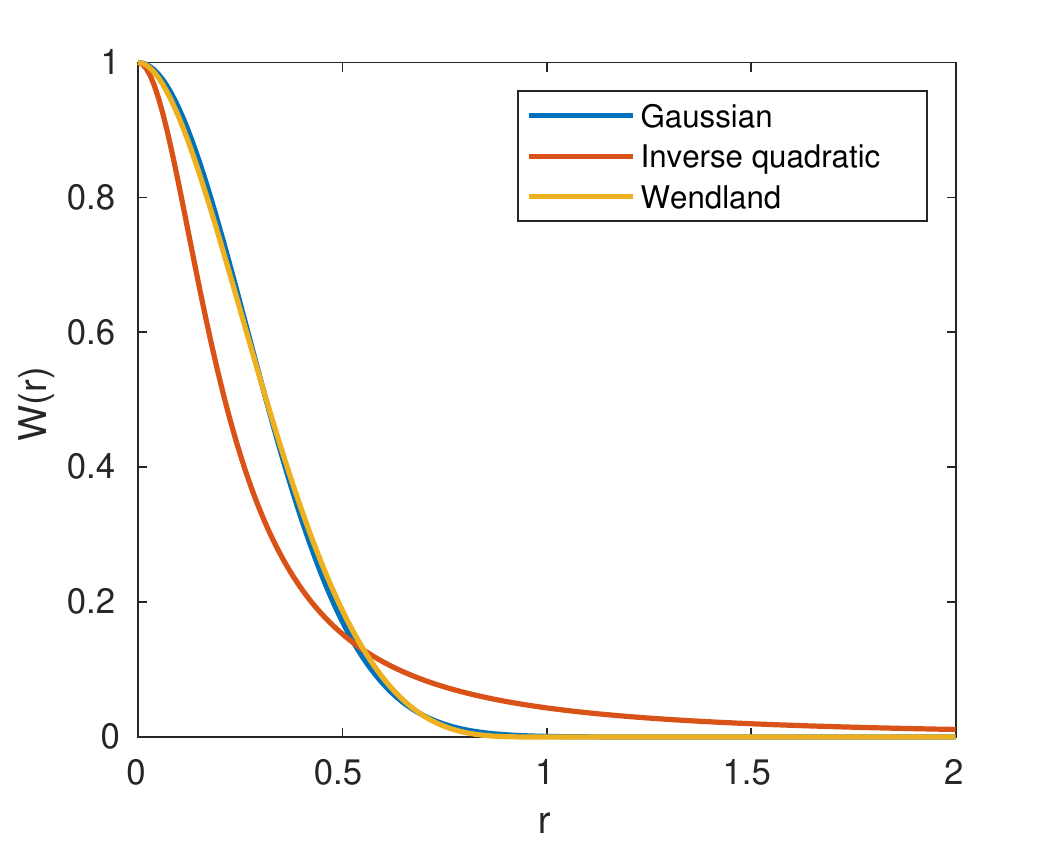}
\caption{\textbf{Radial basis functions.} This graph presents the three considered basis functions after rescaling according to~(\ref{eq:kernel_integral_normalisation}). The Wendland function closely approximates the Gaussian function, yet is compactly supported: it is equal to $0$ for $r \geq 1$~\cite{fornefett2001radial}.}
\label{fig:kernel_comparison}
\end{figure}

To model the spatial correlation between points of the image domain, we define our kernel function as a multi-scale combination of isotropic radial basis functions~\cite{fornefett2001radial} of the form
\begin{equation}
k^{\boldsymbol{\theta}}(\textbf{x},\textbf{x}') = \left[ \sum_{s=1}^S \theta_s \mathcal{K}\left( \frac{\Vert \textbf{x} - \textbf{x}' \Vert}{\rho_s}\right) \right]  \textbf{I}_d,
\label{eq:kernel_bundle}
\end{equation}
where $\textbf{I}_d$ is the $d \times d$ identity matrix, $S$ is the number of scales, and the scale $\rho_s$ of each kernel is set in a pyramidal fashion to $\rho_s = 2^{s-1} \rho_1$. The smallest scale $\rho_1$ is set to $10$ pixels, and the number of scales $S$ is dataset-dependent and defined such that the largest scale approximately matches the size of the image. The function $\mathcal{K} : \mathbb{R}^+ \rightarrow \mathbb{R}^+$ is a nonincreasing radial basis function such that $\mathcal{K}(0) = 1$. We consider three examples of such basis functions in this work: the Gaussian function $\mathcal{K}_G$, the inverse quadratic function $\mathcal{K}_{IQ}$ and the Wendland function $\mathcal{K}_{W}$ of order $1$~\cite{wendland1995piecewise}, respectively defined as
\begin{equation}
\mathcal{K}_G(r) = \exp\left( -\frac{r^2}{r_G^2} \right),
\label{eq:gaussian_kernel}
\end{equation}
\begin{equation}
\mathcal{K}_{IQ}(r) = \frac{1}{1 + \frac{r^2}{r_{IQ}^2}},
\label{eq:inverse_quadratic_kernel}
\end{equation}
\begin{equation}
\mathcal{K}_W(r) = \left( 4 \frac{r}{r_W} + 1\right)  \left( 1 - \frac{r}{r_W} \right)^4_+,
\label{eq:wendland_kernel_1}
\end{equation} where $\left( 1 - r \right)_+  = \max\left(1 - r,0\right)$. Note that the Wendland function is compactly supported and is equal to zero for $r \geq r_W$.
The scaling constants $r_W$, $r_G$ and $r_{IQ}$ encode the spatial scale of each function, playing a similar role as each individual scaling factor $\rho_s$ in our proposed kernel bundle. To ensure a fair comparison between the three basis functions, it is important to set $r_W$, $r_G$ and $r_{IQ}$ so that the three functions correspond to the same scale in the ``default'' setting $\rho_s = 1$. To do so, we arbitrarily set $r_W = 1$ and adjust $r_G$ and $r_{IQ}$ to ensure that 
\begin{equation}
\int_0^{\infty} \mathcal{K}_G(r) \diff r = \int_0^{\infty} \mathcal{K}_{IQ}(r) \diff r = \int_0^{\infty} \mathcal{K}_W(r) \diff r,
\label{eq:kernel_integral_normalisation}
\end{equation}
as suggested in \cite{fornefett2001radial}, leading to $r_G = \frac{2 r_W}{3 \sqrt{\pi}}$ and $r_{IQ} = \frac{2 r_W}{3 \pi}$. A comparison of the three rescaled basis functions is shown in Fig.~\ref{fig:kernel_comparison}. We present an experimental study on the choice of covariance function in Section~\ref{sec:impact_of_chosen_kernel}.

\begin{table*}
\small
\center
\begin{tabular}{|c|c|c|c|c|c|}
\hline
Dataset & Dimensionality & Deformation & Available ground truth & Target set & Annotation uncertainty \\
\hline
CoBrA Lab~\cite{winterburn2013novel} & 3D & Synthetic & All pixels & Brain structure & Fixed \\
CIMA~\cite{borovec2013registration,borovec2018benchmarking,fernandez2002system}  & 2D & Real & Sparse & All pixels & User-specified ellipses \\
Nissl / OCM~\cite{magnain2015optical} & 2D & Real & Dense & All pixels & Fixed \\
COPDgene~\cite{castillo2013reference} & 3D & Real & Dense & All pixels & Fusion of multiple annotations\\
\hline
\end{tabular}
\caption{\textbf{Properties of the four considered datasets.} Each dataset allows the evaluation of some properties of our approach.}
\label{table:dataset_properties}
\end{table*}

\subsubsection{Estimation of the Kernel Parameters}
If no \textit{a priori} knowledge on the transformation is available, we propose to learn the kernel parameters $\boldsymbol{\theta}$ directly on a set  $\mathcal{A}_L$ (more simply called $\mathcal{A}$ in the rest of this section) of $L$ training annotations as follows. We estimate $\boldsymbol{\theta}$ with a predictive approach based on leave-one-out cross validation~\cite{sundararajan2000predictive}. For all $l \in \left\lbrace 1, \ldots, L\right\rbrace $, we denote $\mathcal{A}^{(l)} = \mathcal{A} \setminus \lbrace (\textbf{x}_l, \textbf{y}_l, \Sigma_l) \rbrace$ the set of remaining annotations after removing the $l$-th one. We set $\boldsymbol{\theta}$ as to minimize the negative log-loss equal to 
\begin{equation}
l_{\textrm{GPP}}(\boldsymbol{\theta}) = - \sum_{l=1}^L \log P_{\boldsymbol{\theta}}(\textbf{y}_l \mid \mathcal{A}^{(l)}, \textbf{x}_l, \Sigma_l),
\end{equation}
where $y \mapsto P_{\boldsymbol{\theta}}( y \mid \mathcal{A}^{(l)}, \textbf{x}_l, \Sigma_l)$ is the density of the distribution $\mathcal{N}\left(\mu_{\mid \mathcal{A}^{(l)}}(\textbf{x}_l),k^{\boldsymbol{\theta}}_{\mid \mathcal{A}^{(l)}}(\textbf{x}_l,\textbf{x}_l) + \Sigma_l\right)$. It can be shown~\cite{sundararajan2000predictive} that
\begin{equation}
l_{\textrm{GPP}}(\boldsymbol{\theta}) \propto Ld + \sum_{l=1}^L \left[ \textbf{q}_l(\boldsymbol{\theta})^{\mathrm{T}} D_{ll}(\boldsymbol{\theta})^{-1} \textbf{q}_l(\boldsymbol{\theta}) - \log \det D_{ll}(\boldsymbol{\theta}) \right],
\label{eq:gpp_loss_simplified}
\end{equation}
where $D_{ll}(\boldsymbol{\theta})$ is the $l$-th diagonal $d \times d$ block of $\left(K^{\boldsymbol{\theta}}_{\mathcal{A} \mathcal{A}}\right)^{-1}$, and $\textbf{q}_l(\boldsymbol{\theta})$ is the $l$-th $d \times 1$ segment of the column vector $\textbf{q}(\boldsymbol{\theta}) =  \left( K^{\boldsymbol{\theta}}_{\mathcal{A} \mathcal{A}}\right)^{-1} \left( Y - \mu(\mathcal{X})\right)$.
Equation~(\ref{eq:gpp_loss_simplified}) is the $d$-dimensional generalization of the classical equations for hyperparameter estimation in Gaussian processes~\cite{sundararajan2000predictive}. To minimize $l_{\textrm{GPP}}$, we minimize the right term of~(\ref{eq:gpp_loss_simplified}) using the conjugate gradient method as implemented in the Alglib library~\cite{alglib}.

The presented parameter estimation requires a set of annotations to learn from. For a given application, we assume that one pair of images with landmark correspondences is already available. We preliminarily learn the kernel parameters $\boldsymbol{\theta}$ from this training pair before deploying our model on a new pair. 
 
\section{Experimental Results and Discussion}
\label{sec:experiments}

In this section, we present and discuss experimental results illustrating the advantages of our framework. We implemented our method in C++ and made the code available at \url{https://github.com/LoicPeter/evaluation-deformable-registration}.
All experiments were run on an Intel\textsuperscript{\textregistered} Core\textsuperscript{TM} i7-8650U @ 1.90GHz with 8 cores. In this setup, there is no perceptible waiting time between user interactions whenever the computational simplifications discussed in Section~\ref{sec:efficient_implementation} apply. This is the case for all considered experiments except one (see Section~\ref{sec:target_set_experiments}).

\begin{figure*}[t]
\centering
\captionsetup{position=top}
\subfloat[CoBrA Lab dataset]
{\includegraphics[width=0.23\textwidth]{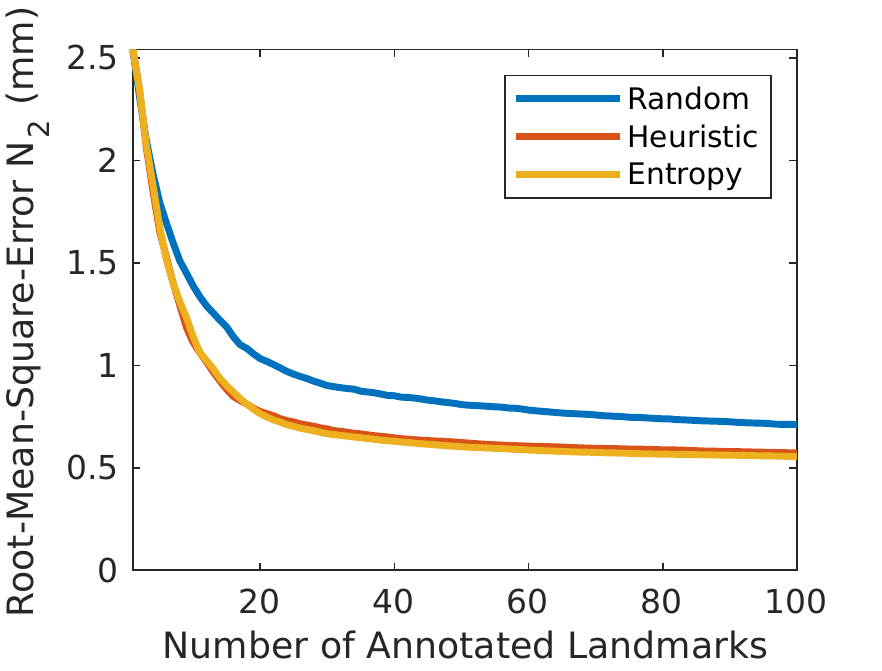}}
\subfloat[CIMA Dataset]{\includegraphics[width=0.23\textwidth]{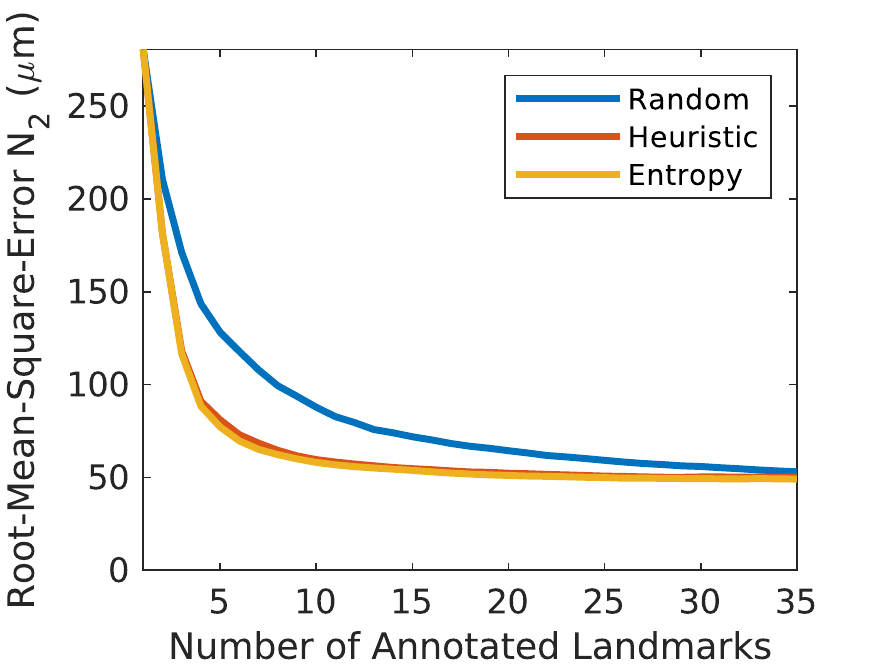}}
\subfloat[Nissl/OCM dataset]{\includegraphics[width=0.23\textwidth]{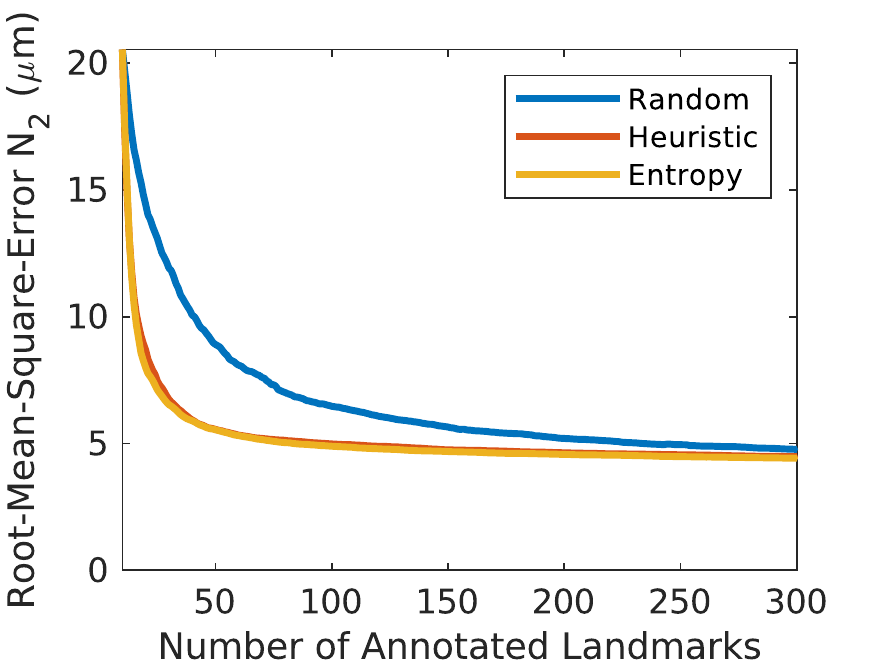}}
\subfloat[COPDgene dataset]{\includegraphics[width=0.23\textwidth]{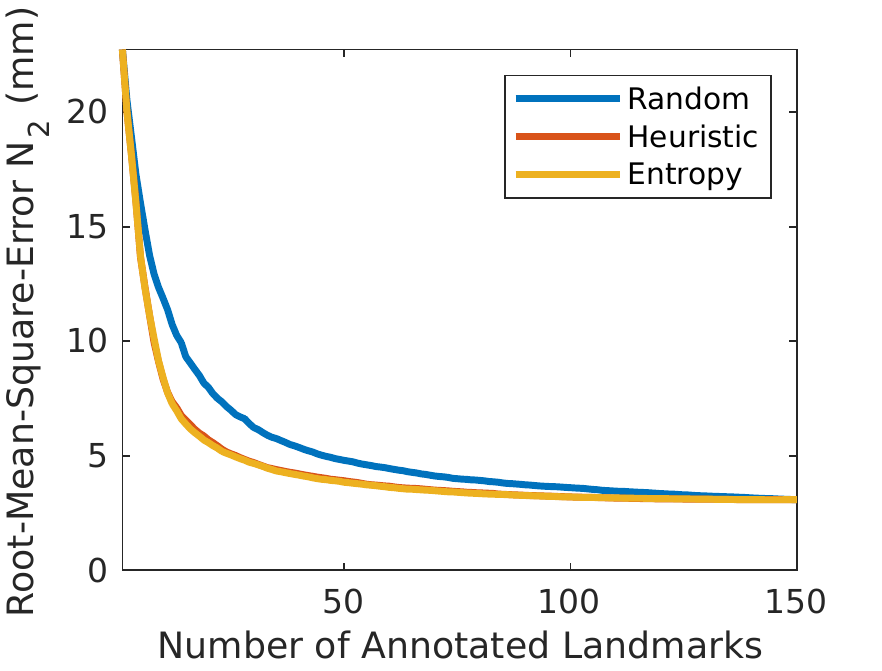}}\hfill
\subfloat
{\includegraphics[width=0.23\textwidth]{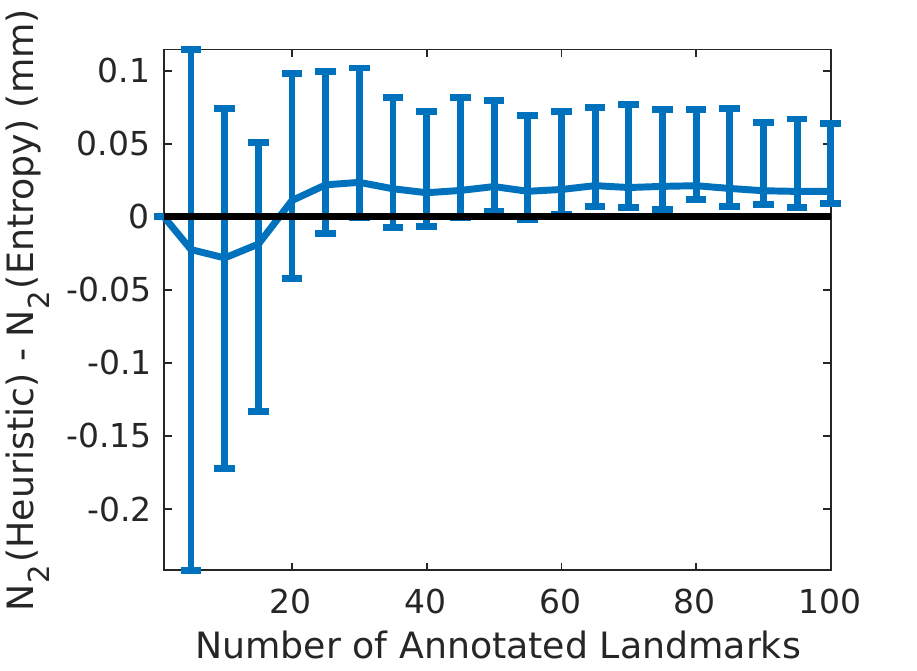}}
\subfloat{\includegraphics[width=0.23\textwidth]{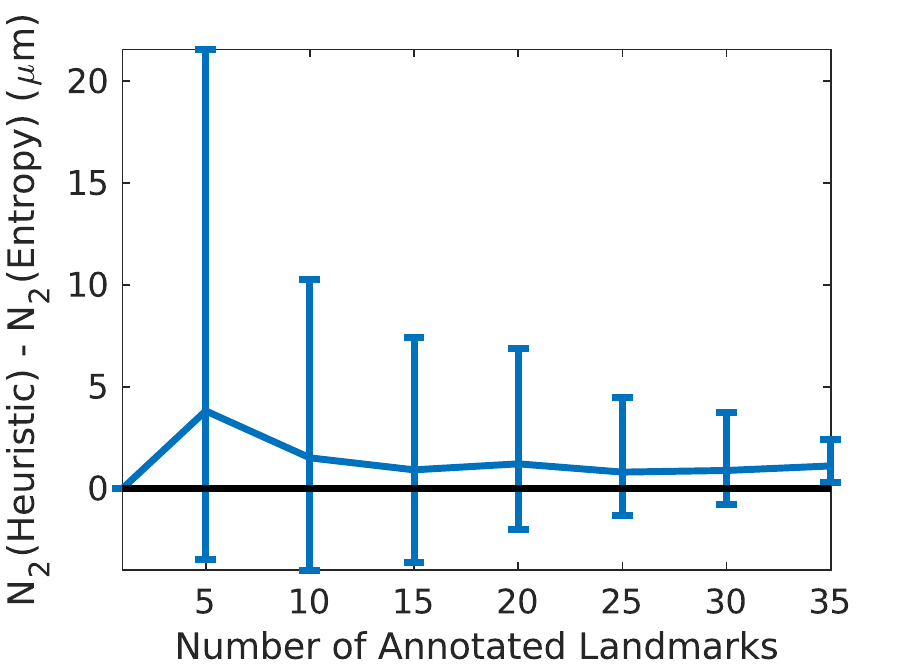}}
\subfloat{\includegraphics[width=0.23\textwidth]{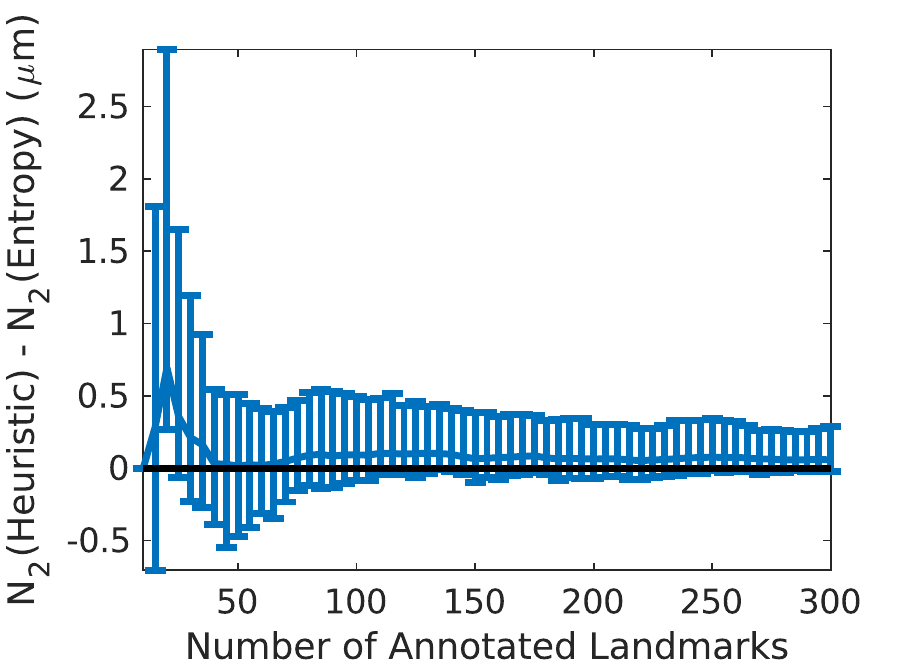}}
\subfloat{\includegraphics[width=0.23\textwidth]{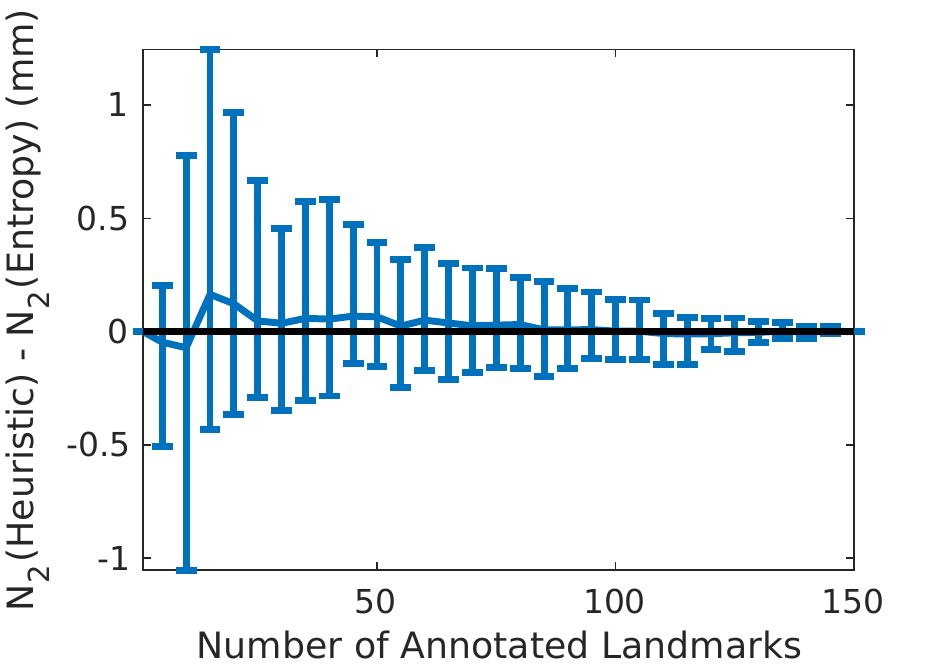}}
\caption{\textbf{Comparison of landmark suggestion strategies in default settings.} We consider here the simplest setting where the whole image domain is set as target and where each landmark annotation is subject to the same fixed annotation uncertainty, without user specification. On the top row, the average root-mean-square error of the mean transformation (as estimated from the Gaussian process model) is reported. On the bottom row, statistics on the difference between the \texttt{Heuristic} suggestion strategy and the proposed \texttt{Entropy} over multiple runs are shown, where the central line is the median difference and the error bars show the first and ninth deciles. A difference over $0$ corresponds to a better performance of \texttt{Entropy} over \texttt{Heuristic}. Across datasets, \texttt{Heuristic} and \texttt{Entropy} clearly outperform \texttt{Random}, and our proposed \texttt{Entropy} strategy performs at least as good as the baseline \texttt{Heuristic}, with a small yet consistent improvement in accuracy of the predicted transformation.}
\label{fig:results_prediction_no_uncertainty}
\end{figure*}

\subsection{Datasets}
\label{sec:datasets}

Our experiments are based on four complementary synthetic and real-world datasets for deformable registration of diverse medical imaging modalities, in 2D and 3D. The considered datasets are summarized in Table~\ref{table:dataset_properties} and described in this section.

\subsubsection{Synthetic T1/T2 Registration (CoBrA Lab Dataset)}

On real data, the non-availability of the true non-linear transformation at all locations of interest, i.e. the very problem that this paper addresses, complicates the experimental validation of our method. To overcome (at least partially) this issue, we first consider a synthetic 3D dataset created from MRI scans made publicly available by the CoBrA Lab~\cite{winterburn2013novel}. This 3D dataset consists of $5$ MRI brain volumes at $\SI{0.6}{\milli\meter}$ resolution, where the T1 and T2 modalities are available and were rigidly registered beforehand. In addition, for each volume, $36$ brain structures were automatically segmented with FreeSurfer~\cite{fischl2012freesurfer} and are used as candidate target sets in our experiments. The set of candidate salient locations $\mathcal{C}$ for annotation was selected with a 3D Harris corner detector. From the provided registered pairs of T1 and T2 images, we create synthetic misaligned pairs for which the true transformation is known by randomly deforming the T2 image (Fig.~\ref{fig:example_t1_t2}) following a similar strategy as in~\cite{iglesias2018joint}: we first randomly deform a coarse 3D grid of control points, which are then linearly interpolated to create a dense stationary velocity field. Finally, we efficiently integrate this field by scaling and squaring~\cite{arsigny2006log}. Note that this synthetic procedure generates transformations that cannot be naturally sampled from a Gaussian process, thereby avoiding an evaluation that would be biased towards our proposed transformation model.

\subsubsection{Intermodal Registration in Histology (CIMA Dataset)}
The publicly available\footnote{Available at:  \url{https://anhir.grand-challenge.org}} CIMA dataset~\cite{borovec2013registration,borovec2018benchmarking,fernandez2002system}  consists of histological sections from $9$ anatomical regions ($3$ lung lesions, $4$ lung lobes and $2$ mammary glands). Five slices were extracted from each region and a different stain was applied to each slice, leading to a total of $9$ inter-modality registration problems with $5$ modalities each. Between $50$ and $100$ landmark correspondences were manually annotated for each region by the authors of the dataset. The image spacing is isotropic and equal to $\SI{3.48}{\micro\meter}$ per pixel. In our context, the two-dimensional nature of this dataset enables the manual placement of ellipses by the user to encode the confidence of each annotated correspondence. To be able to run a large number of experiments on these data, we learned a realistic distribution of the annotation uncertainty by manually reannotating an image pair with confidence ellipses. We then fit a lognormal distribution to the set of the obtained semi-axes to obtain a probability distribution on the ellipse sizes allowing us to draw simulated user annotations in our experiments.

\subsubsection{Nissl/OCM Registration (Nissl/OCM Dataset)} We consider a pair of large 2D
images of Nissl stain (enhancing healthy neurons) and OCM (Optical Coherence Microscopy) acquired on post mortem human brain tissue for which $957$ correspondences
were manually annotated~\cite{magnain2015optical}. Despite the fact that it only consists of a single pair of images, the high density of available gold standard annotations in this dataset allows the evaluation of our framework on a close approximation of a real-world transformation. As this dataset only consists of one pair of images, the kernel parameters cannot be learned on an external annotated pair (see Section~\ref{sec:kernel_choice_and_parameter_estimation}). Instead, we learn the parameters on a subset of $10$ first annotations directly collected on the image pair, where the first annotation is randomly picked and the following ones are sequentially picked as the furthest away from the previously annotated locations (see \texttt{Heuristic} method in Section~\ref{sec:evaluation_suggestion_strategy}).

\subsubsection{Registration of Chest CT Scans (COPDgene Dataset)}
Finally, we consider the publicly available COPDgene dataset\footnote{Available at: \url{https://www.dir-lab.com/Downloads.html}} which comprises $10$ pairs of 3D CT scans corresponding to inspiratory and expiratory breath-hold acquisitions~\cite{castillo2013reference}. $300$ landmark correspondences are available for each of the $10$ pairs. Although the provided correspondences are pointwise, a multi-annotator study was conducted by the authors of the dataset who reported statistics on the inter-user variability on each of the $10$ patients. We exploit these estimates to recreate a multi-expert scenario: for each volume, a mean and standard deviation annotation error are available, so that we simulated three independent experts behaving according to this given error profile, and we merged their annotations to obtain a mean annotation and its sample covariance matrix. Note that, since these uncertainty estimates are obtained by merging several user annotations, it is impractical to obtain them directly at annotation time. Therefore we consider them to be available only at the end of the annotation process: as such, they are not used to guide the suggestion of locations but only to evaluate candidate transformations once the annotation protocol is completed, as described in Section~\ref{sec:uncertainty_aware_mean_square_error}.

\subsection{Evaluation of Landmark Suggestion Strategies}
\label{sec:evaluation_suggestion_strategy}

In this section, we compare the $3$ following strategies for the suggestion of locations to annotate:
\begin{itemize}
\item \texttt{Random}: Queried locations are randomly drawn.
\item \texttt{Heuristic}: Each queried location is iteratively suggested as the furthest from the already annotated ones.
\item \texttt{Entropy}: Our proposed entropy-based suggestions.
\end{itemize}
To quantify the informativeness of a suggestion strategy, we measure at each iteration the quality of the mean transformation $\mu_{\mid \mathcal{A}}$ predicted by our Gaussian process given the acquired annotations. The similarity between $\mu_{\mid \mathcal{A}}$ and $\phi$ is quantified by a norm $\Vert \Delta_{\mathcal{T}}(\phi,\mu_{\mid \mathcal{A}})\Vert_p$ over the target set $\mathcal{T}$, as defined in~(\ref{eq:p_norms}). In the presented results, we refer to these norms as $N_p$ for compactness. As our experiments consist of multiple runs on a same dataset, the results are presented in a compact way as follows. First, we report the mean curve over runs. In addition, to get better insights on the difference between the two most competitive approaches $\texttt{Heuristic}$ and $\texttt{Entropy}$, we report the statistical distribution of the difference between the respective results obtained with each approach. The median difference is displayed as central line and bars represent the first and ninth deciles. A difference over $0$ (shown as a black line) indicates a better performance of $\texttt{Entropy}$ over $\texttt{Heuristic}$. The three following experimental settings were investigated.

\subsubsection{Fixed Uncertainty and $\mathcal{T} = \Omega$}.
\label{sec:default_experimental_setting}

We conducted on the four datasets a first series of experiments in what can be seen as a ``default'' setting: all pixels of the image domain are taken as target, and each landmark is associated to a fixed and small isotropic uncertainty (i.e., not given by the annotator). We used half of the available landmarks as annotable locations, and the remaining half as an evaluation set on which the accuracy of the mean transformation is estimated. The results are presented in Fig.~\ref{fig:results_prediction_no_uncertainty}, where the experiments are repeated over multiple runs to obtain at least $100$ result curves on each dataset. The results demonstrate that \texttt{Heuristic} and \texttt{Entropy} display similar performance in average and clearly outperform a random selection of landmark locations (Fig.~\ref{fig:results_prediction_no_uncertainty}). This is intuitively explained by the fact that, in the aforementioned scenario, an optimal strategy consists in picking landmarks covering the spatial domain as uniformly as possible: this is by construction what the \texttt{Heuristic} does, and this strategy is naturally recovered with our entropy-based approach. In contrast, a random selection may result in undersampling locations in some areas. Moreover, once the number of annotations is large enough to reach convergence, a small yet consistent improvement is obtained with our proposed \texttt{Entropy} approach, as mostly apparent on the statistical distribution of the difference between their respective errors. 

\subsubsection{Specification of a Target Set $\mathcal{T}$}
\label{sec:target_set_experiments}
\begin{figure}[t]
\centering
\captionsetup[subfigure]{position=top}
\subfloat
{\includegraphics[width=0.23\textwidth]{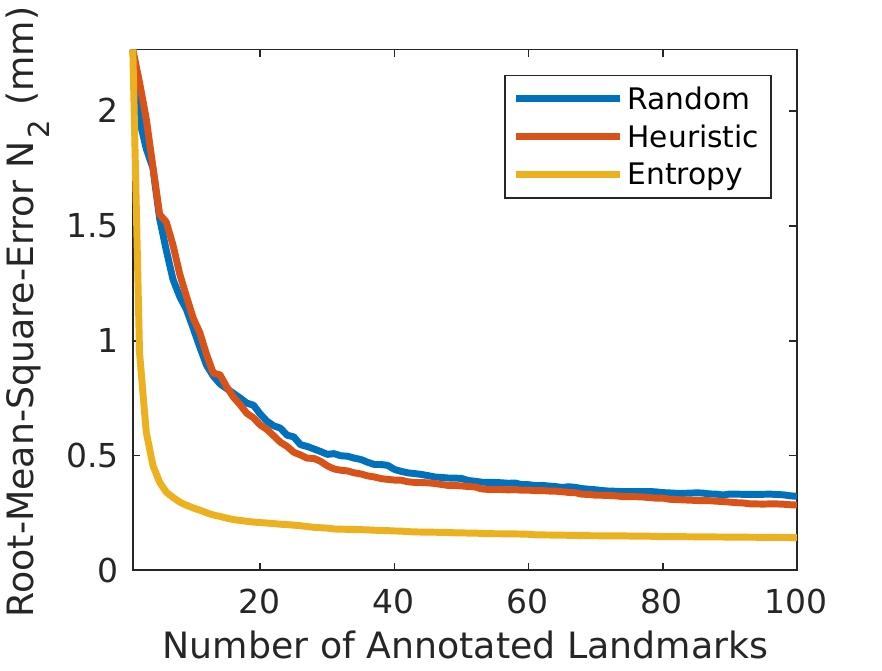} }
\subfloat{\includegraphics[width=0.23\textwidth]{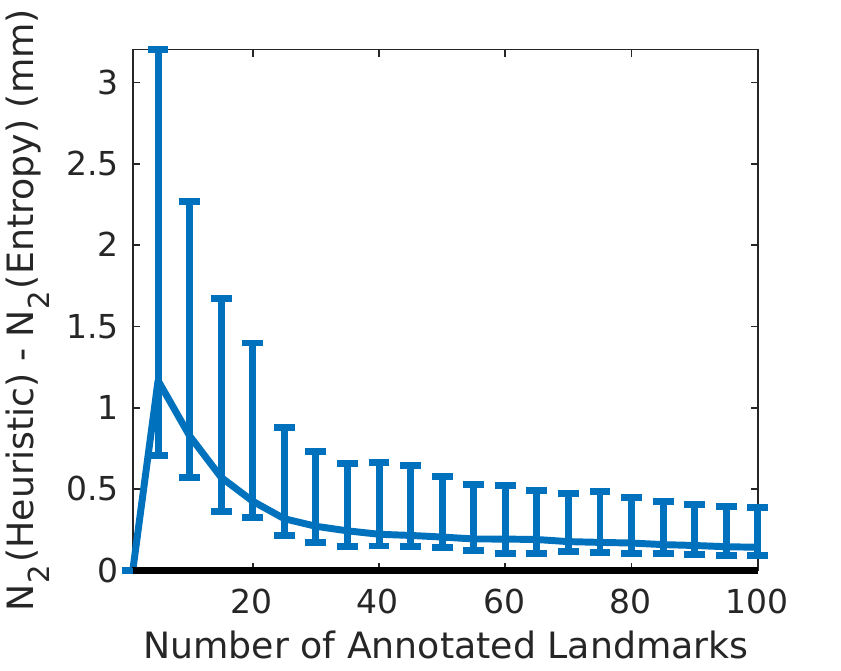} }\hfill
\caption{\textbf{Annotations informing on a given target structure.} This figure shows experimental results for multiple runs on the CoBrA Lab dataset, where the target $\mathcal{T}$ is extracted from a different brain structure at each run. In this scenario, the results show that our landmark suggestion approach \texttt{Entropy} yields the smallest error and the fastest convergence by focusing on the areas of the image domain that are relevant to the target.}
\label{fig:results_brain_dataset_prediction}
\end{figure}

On the CoBrA Lab dataset, we investigated the effect of choosing a specific target area $\mathcal{T}$. We run experiments on $10$ image pairs generated from the $5$ available brains (with $2$ randomly drawn synthetic deformations per brain). For each pair, $10$ random candidate brain structures were consecutively considered among the available segmentation labels. In each case, we extracted the target samples $\mathcal{T}$ from the edge of the considered structure. As in the first experimental setting, we simulated a user providing annotations  of fixed isotropic uncertainty: all ellipses are circles of radius $1$ voxel = \SI{0.6}{\milli\meter}. The results presented in Fig.~\ref{fig:results_brain_dataset_prediction} show that \texttt{Entropy} yields a clear improvement by encouraging suggestions that are informative with respect to the target structure. This study shows that our approach can be, if relevant for the considered application, tailored to a subset of the image domain in a mathematically principled way.

\subsubsection{Annotations with User-Specified Uncertainty}
\label{sec:evaluation_uncertain_annotations_prediction}

\begin{figure}[t]
\centering
\captionsetup[subfigure]{position=top}
\subfloat[Normal uncertainty\label{fig:prediction_cima_normal_uncertainty}]
{\includegraphics[width=0.23\textwidth]{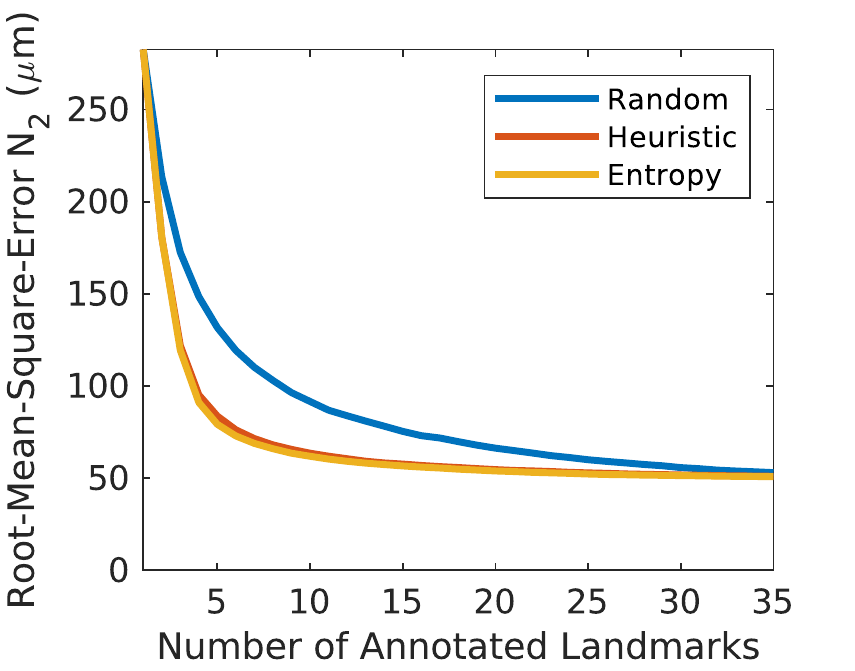} }
\subfloat[Five-fold increase in uncertainty\label{fig:prediction_cima_fivefold_uncertainty}]{\includegraphics[width=0.23\textwidth]{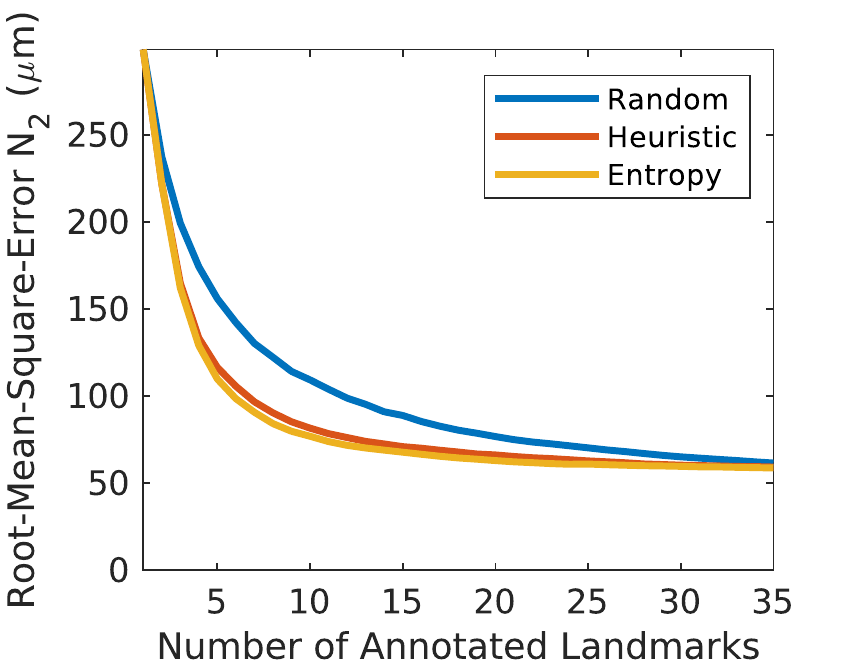} }\hfill
\subfloat{\includegraphics[width=0.23\textwidth]{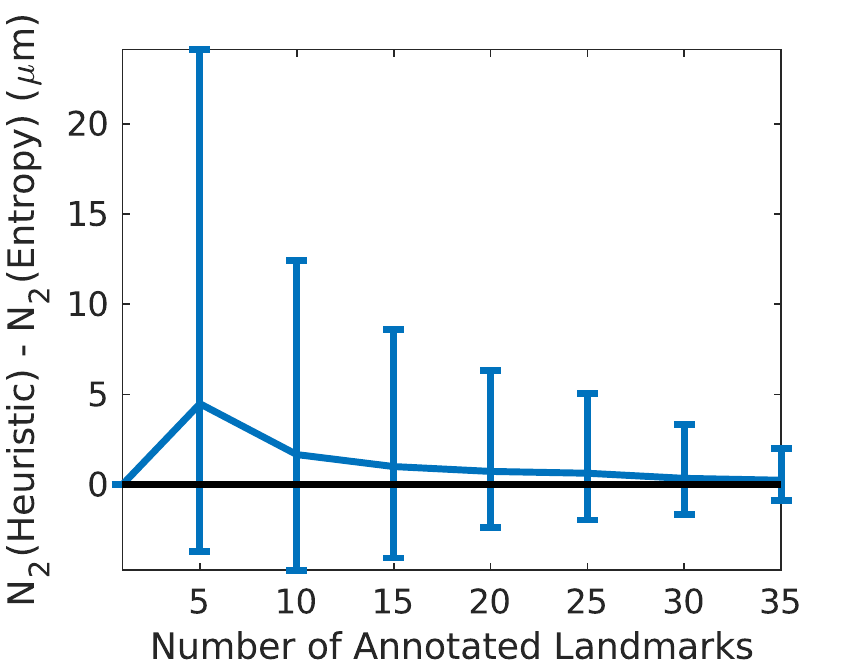}}
\subfloat{\includegraphics[width=0.23\textwidth]{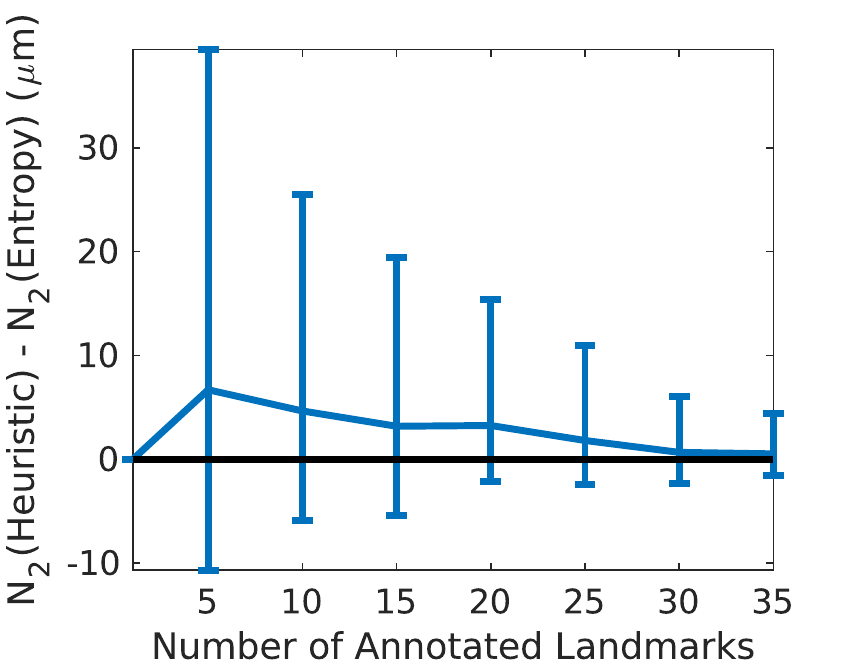}}\hfill
\caption{\textbf{Annotations with user-specified confidence ellipses.} (a) On the CIMA dataset where user-specified elliptic uncertainties are considered, \texttt{Entropy} yields a small but consistent improvement over \texttt{Heuristic}. (b) Results on a modified setup where the range of uncertainties provided by the annotator is artificially increased, which shows a correlation between the size of the observed improvement and the range of uncertainties.}
\label{fig:results_cima_dataset_prediction}
\end{figure}

We studied, on the CIMA dataset for which uncertainties can easily be communicated at annotation time, the effect of specifying an uncertainty ellipse with each annotation in comparison to simpler pointwise correspondences. We considered a simulated user who provides, at each queried location, an ellipse whose two semi-axes are randomly drawn from the learned annotation distribution (Section~\ref{sec:datasets}). The experimental setup otherwise follows the default setting presented in Section~\ref{sec:default_experimental_setting}. The results are reported in Fig.~\ref{fig:prediction_cima_normal_uncertainty} and show a small improvement similar to the one observed  in Fig.~\ref{fig:results_prediction_no_uncertainty} with fixed uncertainty. 

We then run again the same experiment where, this time, we artificially applied an increase on the range of the provided ellipse sizes by applying a fivefold scaling on the learned distribution of annotations. The results show that, in this setup where the range of provided uncertainties is higher, the improvement brought by \texttt{Entropy} over the \texttt{Heuristic} increases (Fig.~\ref{fig:prediction_cima_fivefold_uncertainty}). This behavior is explained by the fact that an entropy-based strategy is able to focus on areas where the informativeness of the previous annotations was lower. In contrast, the position-based heuristic approach solely considers the distance to the already provided annotations, regardless of the confidence of the user on each annotation.

\begin{figure*}[t]
\centering
\captionsetup{position=top}
\subfloat[CoBrA Lab dataset]
{\includegraphics[width=0.23\textwidth]{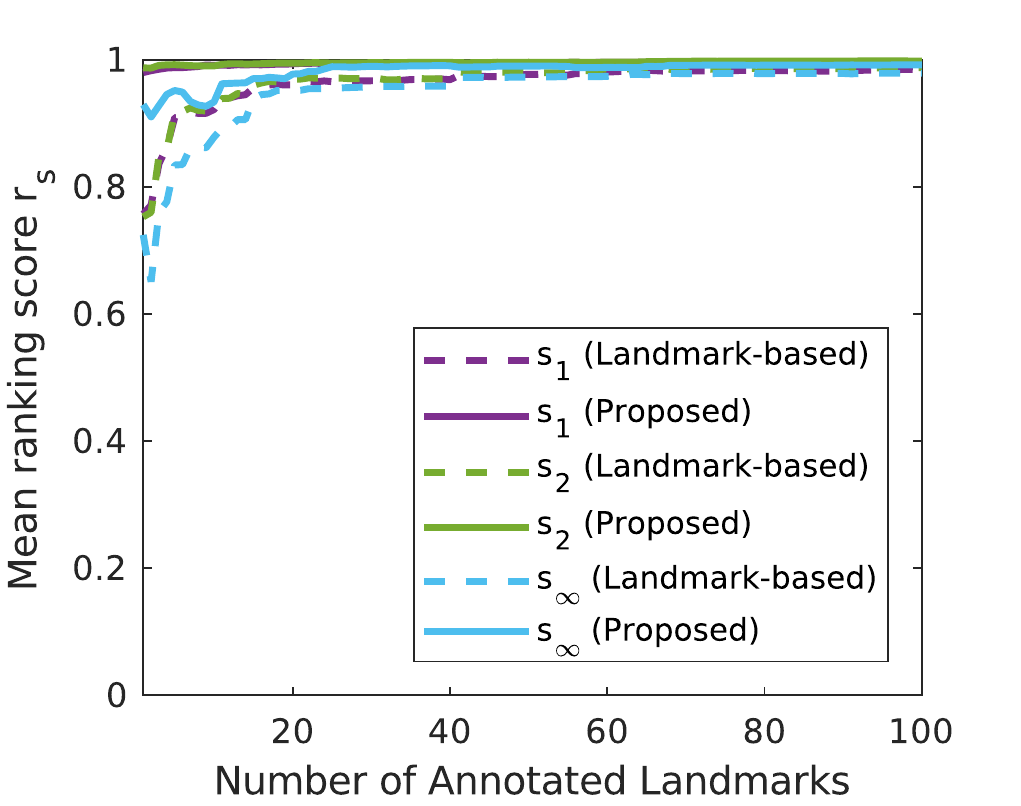}}
\subfloat[CIMA Dataset]{\includegraphics[width=0.23\textwidth]{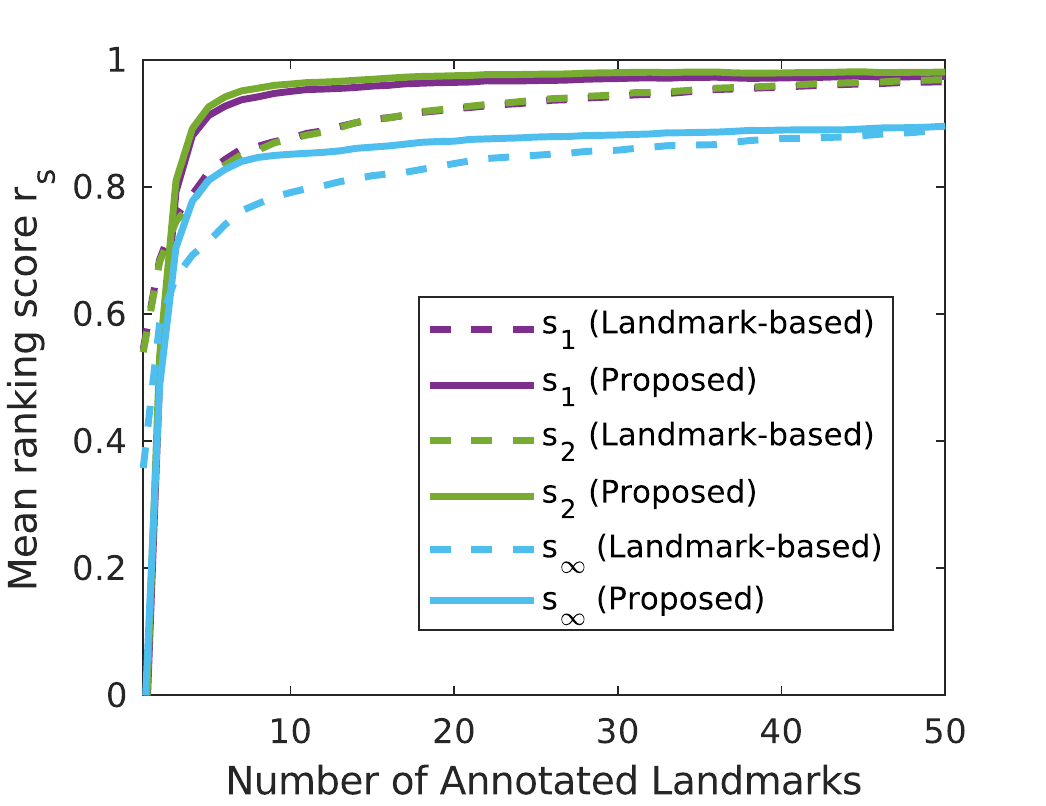}}
\subfloat[Nissl/OCM dataset]{\includegraphics[width=0.23\textwidth]{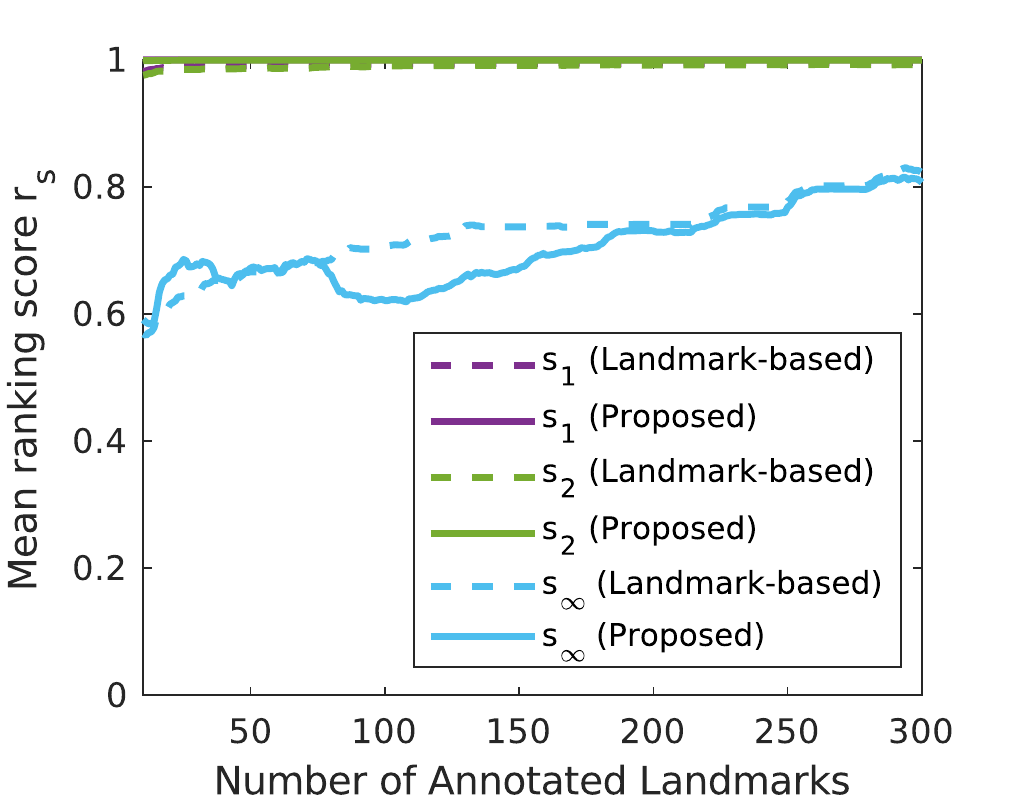}}
\subfloat[COPDgene dataset]{\includegraphics[width=0.23\textwidth]{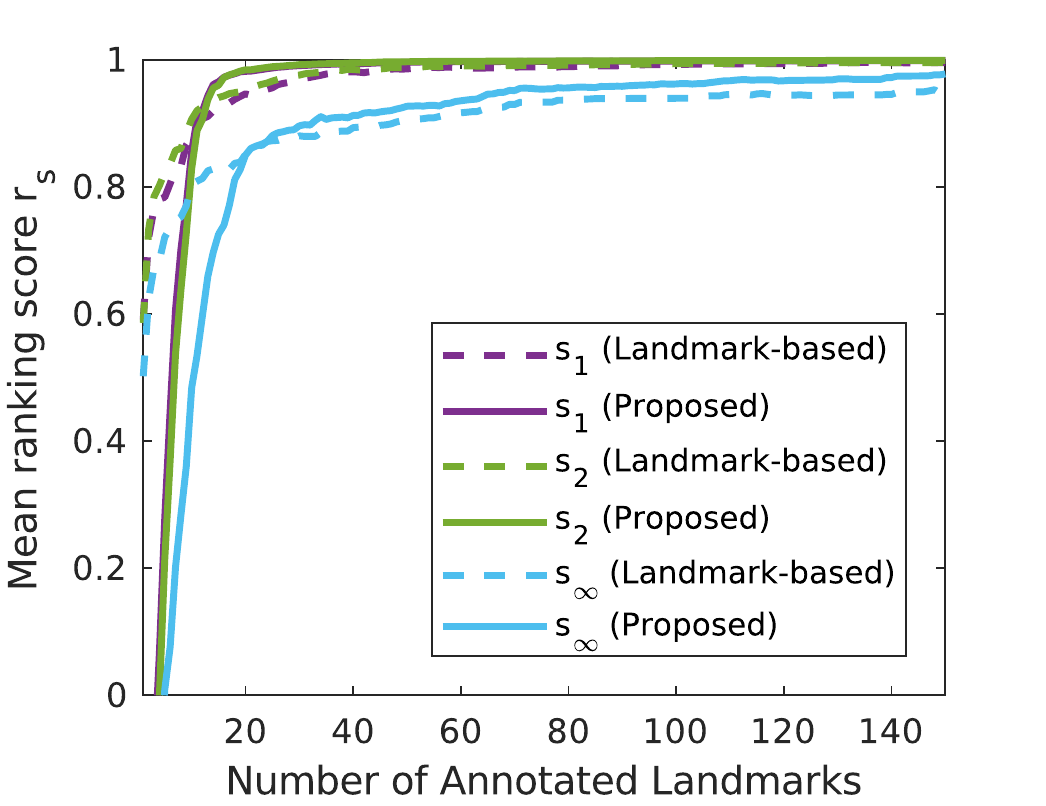}}\hfill
\subfloat
{\includegraphics[width=0.23\textwidth]{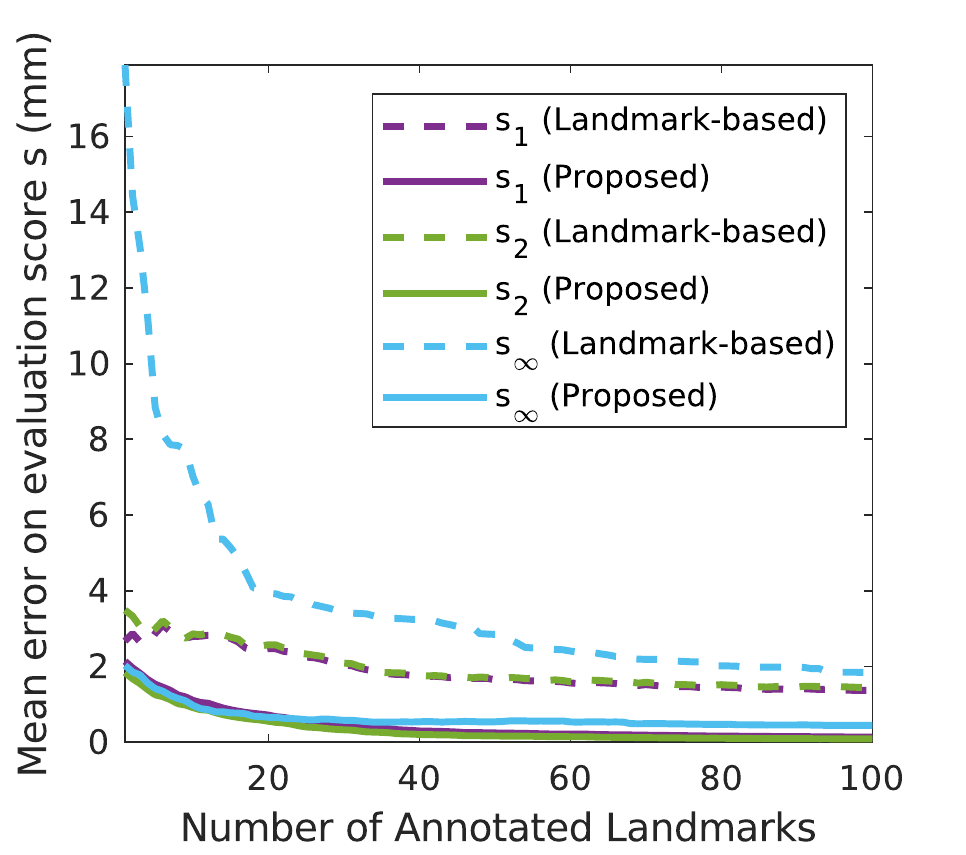}}
\subfloat{\includegraphics[width=0.23\textwidth]{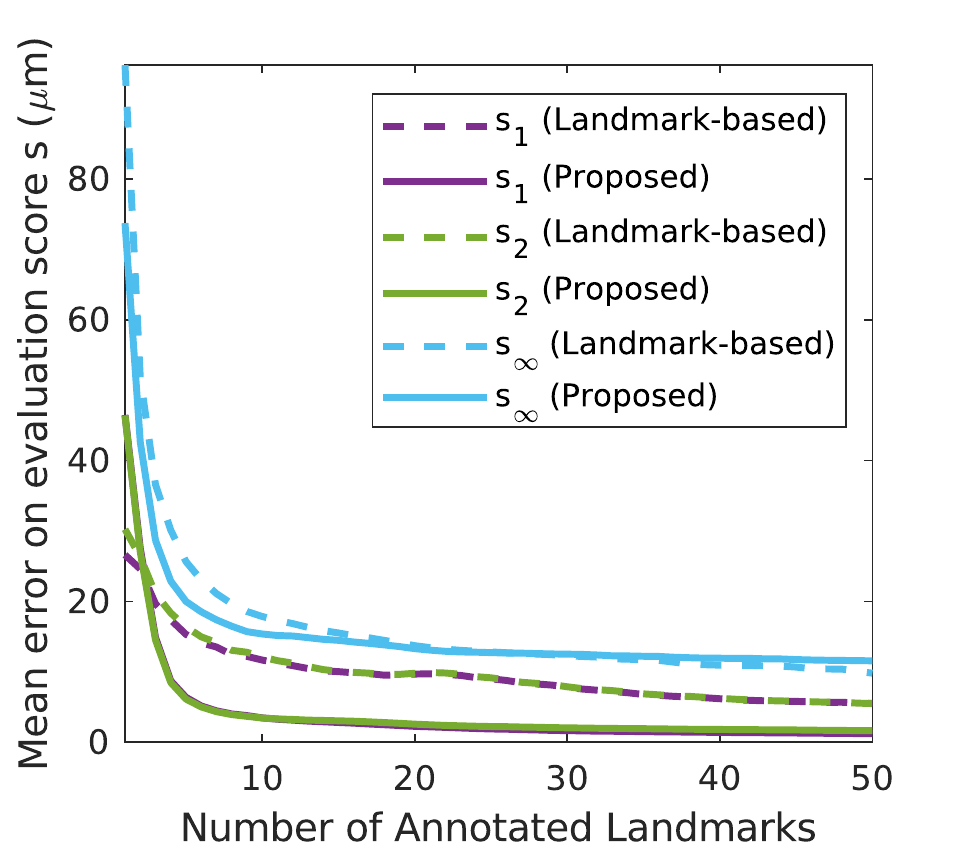}}
\subfloat{\includegraphics[width=0.23\textwidth]{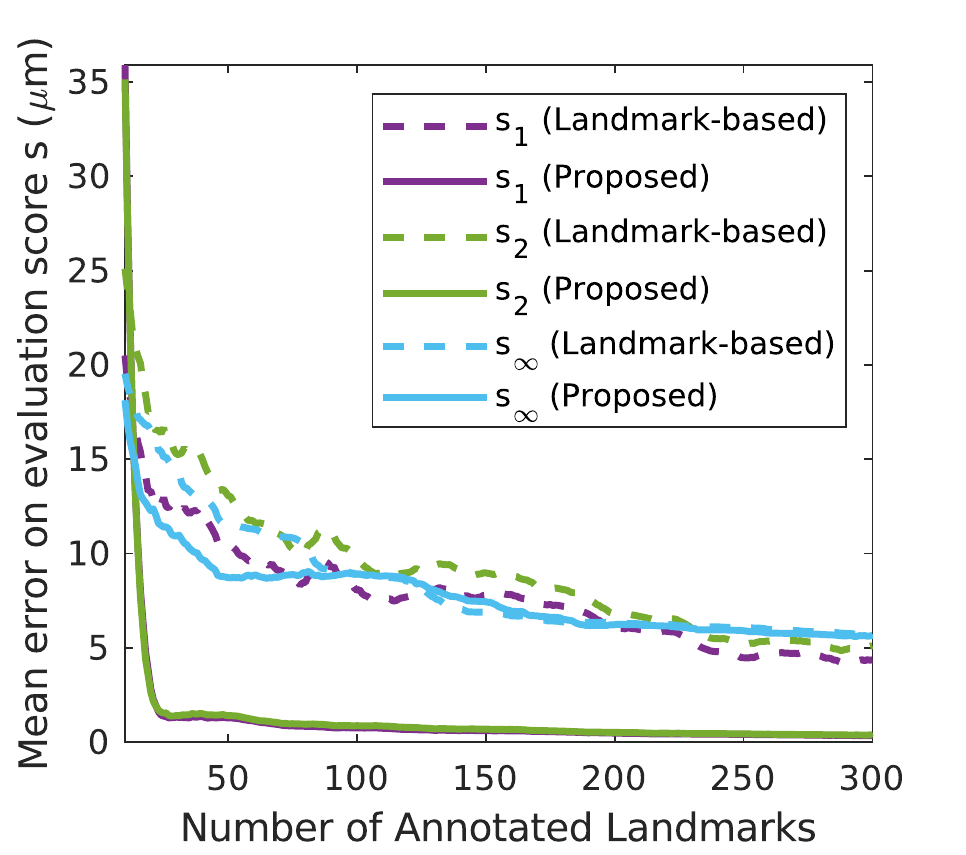}}
\subfloat{\includegraphics[width=0.23\textwidth]{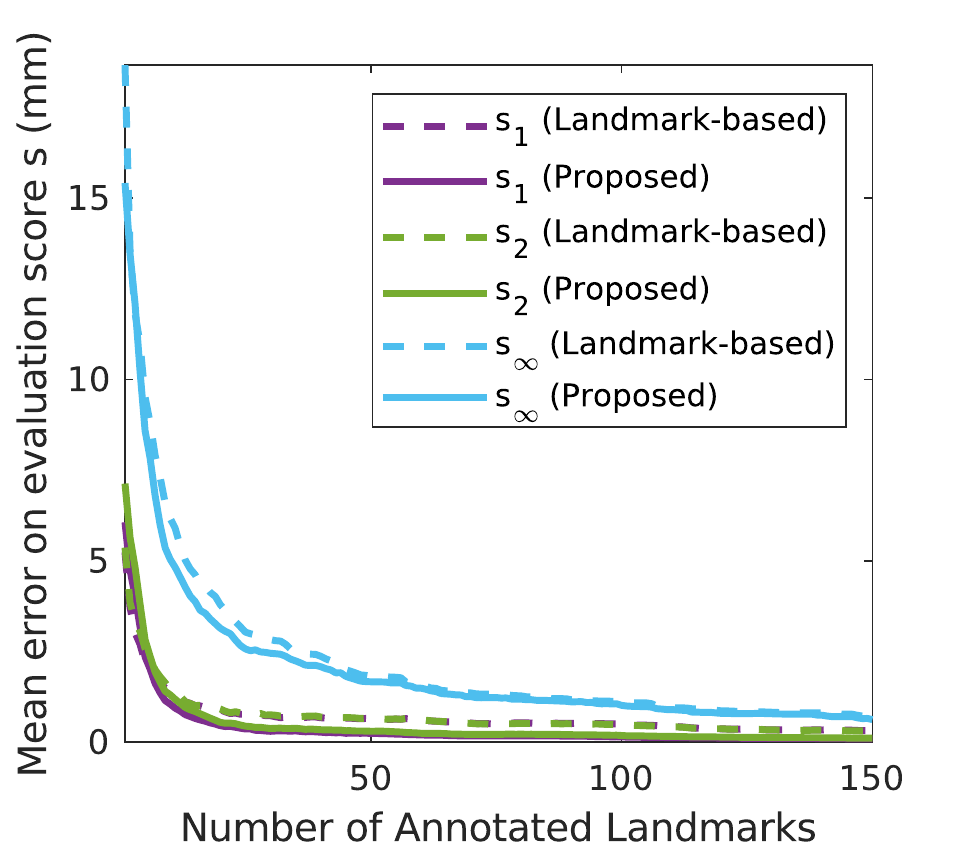}}
\caption{\textbf{Evaluation of candidate transformations obtained from registration algorithms.} We experimentally measured the ability of our proposed method to assess the quality of the output of registration algorithms and compared it with a standard evaluation method based on landmark annotation alone. We use two measures to quantify the evaluation capacity of a method: a correlation score stating how accurately candidate transformations are ranked (top row), and a standard mean estimate of the spatial error (bottom row). For each experiment, several target evaluation metrics ($1$-norm, $2$-norm and $\infty$-norm) were considered.  Our uncertainty-aware evaluation strategy overall improves over the classical landmark-based approach.}
\label{fig:results_evaluation_capacity}
\end{figure*}

\subsection{Ability to Evaluate Candidate Transformations}
\label{sec:experiments_capacity_evaluation}

In the previous section, we compared suggestion strategies by measuring the accuracy of the mean prediction of the Gaussian process when conditioned on the queried annotations. To assess the capacity of our Gaussian process model to evaluate registration algorithms, we conducted on the four datasets a set of experiments where, for any given pair to register, candidate estimated transformations $\hat{\phi}_1, \ldots, \hat{\phi}_K$  were first obtained as the output of $K$ different registration methods. We then measured the ability of our method to estimate the quality of each of the obtained transformations, as desired in an evaluation context. For each pair, the transformations to rank were obtained using the Elastix toolbox~\cite{klein2009elastix,shamonin2014fast}, each candidate transformation corresponding to a different set of registration parameters. For all datasets, we investigated two number of iterations ($500$ and $1000$), five number of resolutions (from $4$ to $8$) and three optimization metrics: Mattes mutual information, normalized mutual information and normalized correlation. Given its intra-modality nature, we also considered the mean-square metric for the COPDgene dataset. All other registration parameters were kept constant: for more details, the complete Elastix parameter files can be found in the supplementary material. Considering all combinations of parameters, we obtained $K = 30$ candidate transformations for each image pair ($K = 40$ for the COPDgene dataset). 

Given the true transformation $\phi$ relating two images, we denote $s_p(\hat{\phi}) = \Vert \Delta_{\mathcal{T}}(\phi,\hat{\phi}) \Vert_p$ the true (unknown in a real-world scenario) score of a candidate transformation $\hat{\phi}$, i.e. the mean error (in a $L^p$ sense) between the true and estimated transformations. By ordering the scores $s_p(\hat{\phi}_1), \ldots, s_p(\hat{\phi}_K)$, a rank $\rk_{\textrm{true}}(\hat{\phi}_k)  \in \left\lbrace 1, \ldots, K\right\rbrace$ of each $\hat{\phi}_k$ is defined. We measure the evaluation capacity of a method via its ability to:
\begin{enumerate}
\item predict the correct rank of each registration result $\hat{\phi}_k$;
\item predict the correct error score $s_p(\hat{\phi}_k)$.
\end{enumerate}
We quantitatively measure the similarity between the predicted and true rankings using Spearman's rank correlation coefficient
\begin{equation}
r_s = 1 - \frac{6}{K \left( K^2 - 1\right)} \sum_{k=1}^K \left(\rk_{\textrm{pred}}(\hat{\phi}_k) - \rk_{\textrm{true}}(\hat{\phi}_k)\right)^2,
\end{equation}
which satisfies $r_s \in \left[-1,1\right]$, where $1$ indicates a perfect ranking. The accuracy of the prediction of $s_p$ is classically measured using the mean absolute error between the predicted and true scores given in units of length (here, in \SI{}{\milli\meter} or \SI{}{\micro\meter}).

These two measures, although naturally related, enable complementary interpretations of the results: the prediction of the score of each transformation is important to interpret quantitatively the error made by each transformation in spatial units (e.g. to know whether the best transformation is of acceptable accuracy for the application at hand), while the ranking measure gives an indication on the degree of accuracy needed to differentiate the quality of the outputs naturally obtained from different registration algorithms.

In this context, we compare the two following strategies:
\begin{itemize}
\item \texttt{Landmark-based}: The standard landmark-based evaluation~(\ref{eq:set_landmark_displacements}), where the true score $s_p(\hat{\phi}_k)$ is approximated as the mean error $\Vert \Delta_{\mathcal{L}}(\phi,\hat{\phi})\Vert_p$ over the set of user-provided landmarks $\mathcal{L}$, without accounting for any user uncertainty;
\item \texttt{Proposed}: The score $s_p(\hat{\phi}_k)$ is estimated via the mean prediction of the Gaussian process over the target set $\mathcal{T}$, i.e. as $\Vert \Delta_{\mathcal{T}}(\phi,\hat{\phi} \mid \mathcal{A}) \Vert_p$ (see Section~\ref{sec:uncertainty_aware_mean_square_error}).
\end{itemize}

We report in Fig.~\ref{fig:results_evaluation_capacity} the mean results on the four datasets, where we considered different ``true'' scores $s_p$ for $p = 1, 2, \infty$. For these experiments, the target set was defined as the whole image domain. The presented results are aggregated over $10$ runs. Our approach consistently compares favorably to the landmark-based one over the four datasets, and requires fewer annotations before converging. The complementary nature of our two considered metrics is apparent on the Nissl/OCM dataset: although both landmark-based and our proposed evaluation method are able to rank very accurately the candidate transformations obtained from multiple registration algorithms, our method is much more effective to accurately assess the spatial error made by these estimated transformations. The results also demonstrate that choosing $p = 1$ or $p = 2$ as evaluation score yield nearly identical evaluation results. In contrast, the results obtained for $p = \infty$ are noisier, which can be explained by their outlier-sensitive nature. Indeed, the score $s_\infty(\hat{\phi})$ is defined by a single target location, namely the one where the error made by the transformation $\hat{\phi}$ is the highest. 

Several factors may explain the observed difference in evaluation performance between the two approaches. First, our approach takes into account the uncertainty on each annotation when available (CIMA and COPDgene datasets) to estimate the underlying true transformation. As some annotations may be subject to high inaccuracies (either encoded as a large ellipse or estimated by a high disagreement between multiple experts), it is benefical to downweigh these noisier annotations. We can notice that the overall improvement is stronger on the CIMA dataset than on the COPDgene dataset, which can be attributed to the dataset properties. The annotation uncertainties are generally higher on the CIMA dataset, whereas the inter-user variability on the COPDgene annotations is substantially smaller (at most a difference of a few voxels at each location).

Moreover, our approach estimates the transformation score based on all the locations of interest defined in the set $\mathcal{T}$ via the mean Gaussian process prediction at these locations, which more accurately reflects the areas that may be underrepresented in the provided landmark annotations. This property, together with the local uncertainty estimated by the Gaussian process at each target location, is also what enables a qualitative visualization of statistical errors on the entire image domain by means of a statistical heat map, an example of which is shown in Fig.~\ref{fig:example_heat_map}.

Further on the effect of the chosen $L^p$ norm, we can note that our proposed evaluation score (\ref{eq:target_set_displacements_on_annotations}) is based on the \textit{mean} Gaussian process prediction only. In particular, although this mean prediction accounts for the uncertainty associated to each provided annotation, the confidence of the Gaussian process prediction itself is not taken into account in our evaluation strategy. Although this can be seen as a limitation, in the $L^2$ case, it can be shown (see e.g.~\cite{sundararajan2000predictive}) that:
\begin{multline}
\mathbb{E}_{\phi \mid \mathcal{A}}\left[ \Vert \Delta_{\mathcal{T}}(\phi,\hat{\phi}) \Vert_2^2 \right] = \Vert \Delta_{\mathcal{T}}(\phi,\hat{\phi} \mid \mathcal{A}) \Vert_2^2\\
+ \sum_{\textbf{x} \in \mathcal{T}} \tr\left( k_{\mid \mathcal{A}}(\textbf{x},\textbf{x})\right),
\end{multline}
where the sum of traces remarkably does not depend on $\hat{\phi}$. 
In other words, if the root-mean-square-error $s_2$ is used as evaluation metric, our proposed ranking which measures the deviation with respect to the mean transformation $\mu_{\mid \mathcal{A}}$ is equivalent to ranking with respect to the (uncertainty-aware) expected mean square error between $\phi$ and $\hat{\phi}$. Although statistics on~(\ref{eq:target_set_displacements_on_annotations}) only approximate the full posterior distribution in the case of other metrics, other approaches can be considered due to the availability of a probabilistic model. For example, Monte Carlo estimations could be conducted based on transformations sampled from the distribution conditioned on $\mathcal{A}$.

\subsection{Effect of the Chosen Kernel Function}
\label{sec:impact_of_chosen_kernel}

The presented probabilistic model relies on a kernel function which we defined as a multi-scale bundle of Wendland basis functions (Section.~\ref{sec:kernel_choice_and_parameter_estimation}). We explored the sensitivity of our model to the chosen kernel function and specifically addressed the influence of the chosen radial basis function and the effect of choosing a multi-scale bundle. The experiments were run on the CIMA and the COPDgene datasets for which many images or annotations are available (with respectively 2D and 3D kernels), in the ``default'' experimental setting of Section~\ref{sec:evaluation_suggestion_strategy}.

\subsubsection{Comparison of Basis Functions}

\begin{figure}[t]
\centering
\captionsetup{position=top}
\subfloat[CIMA dataset]
{\includegraphics[width=0.23\textwidth]{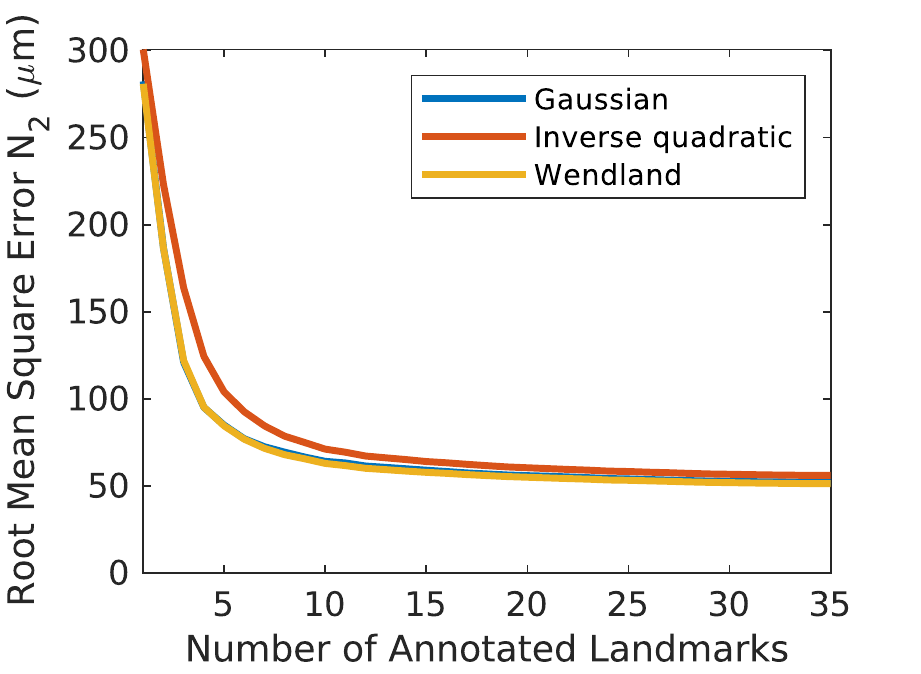}}
\subfloat[COPDgene dataset]{\includegraphics[width=0.23\textwidth]{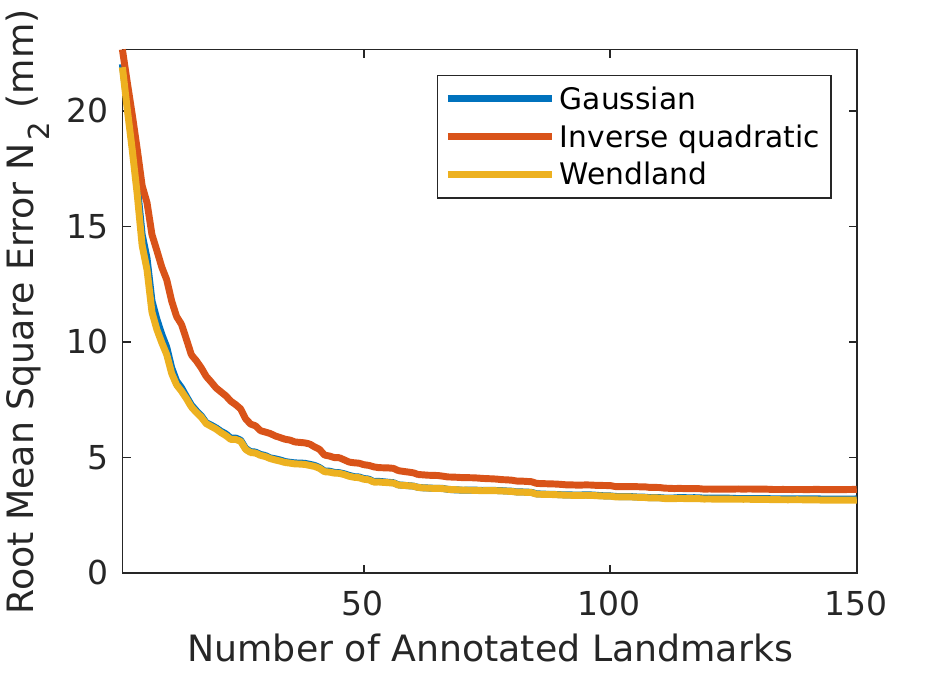}}\hfill
\subfloat{\includegraphics[width=0.23\textwidth]{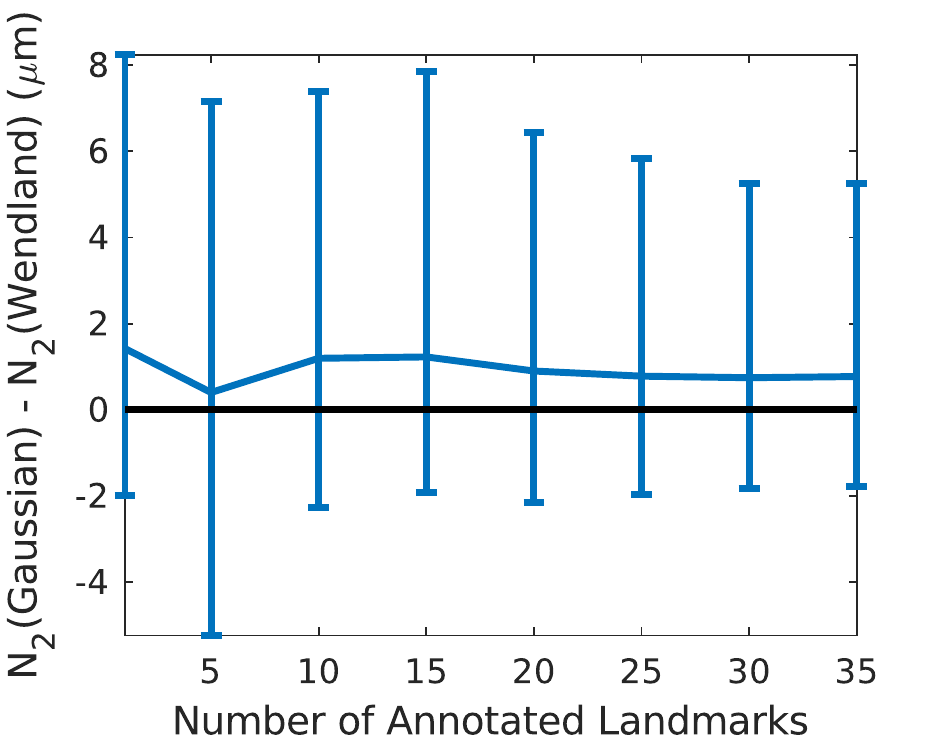}}
\subfloat{\includegraphics[width=0.23\textwidth]{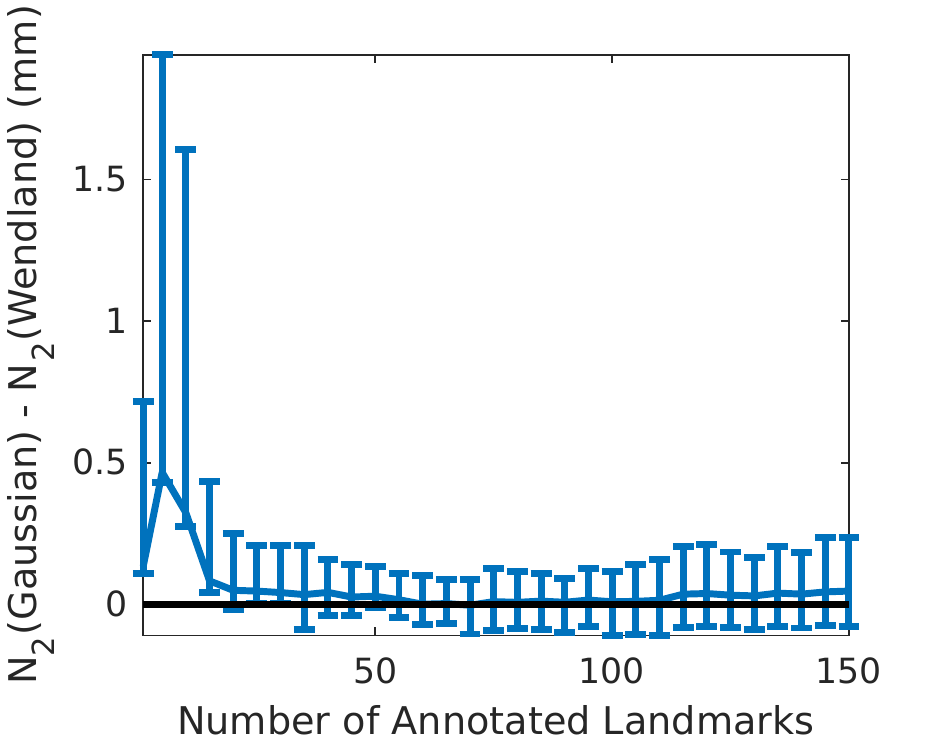}}\hfill
\subfloat{\includegraphics[width=0.23\textwidth]{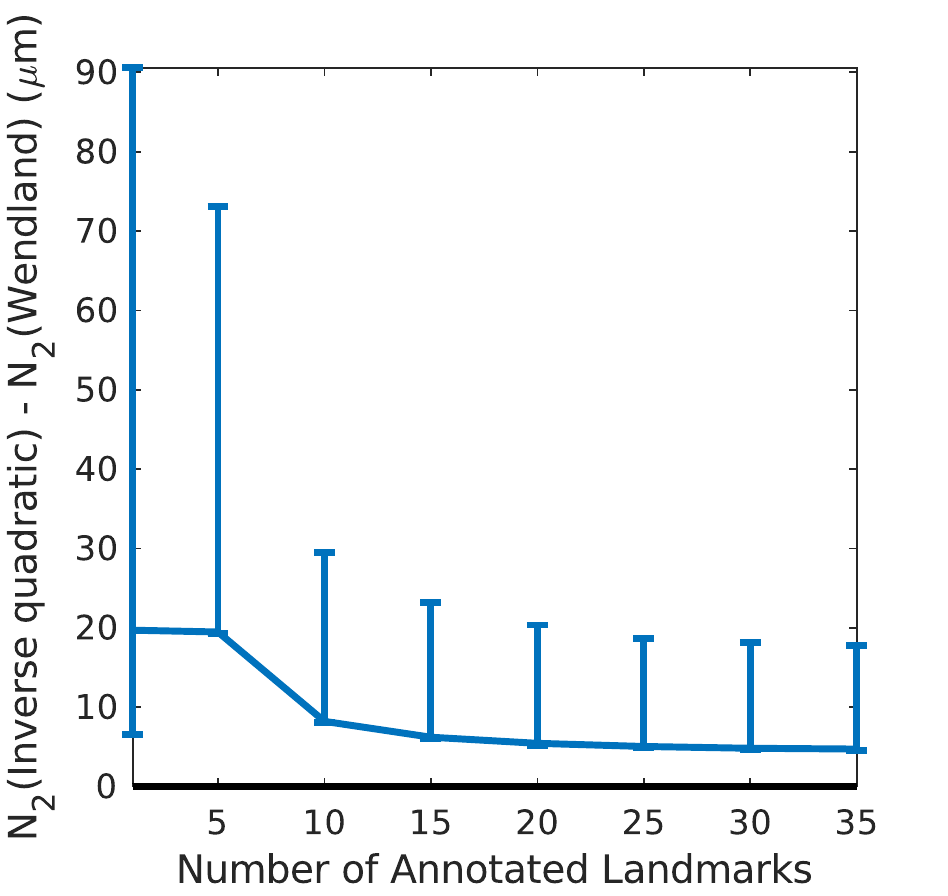}}
\subfloat{\includegraphics[width=0.23\textwidth]{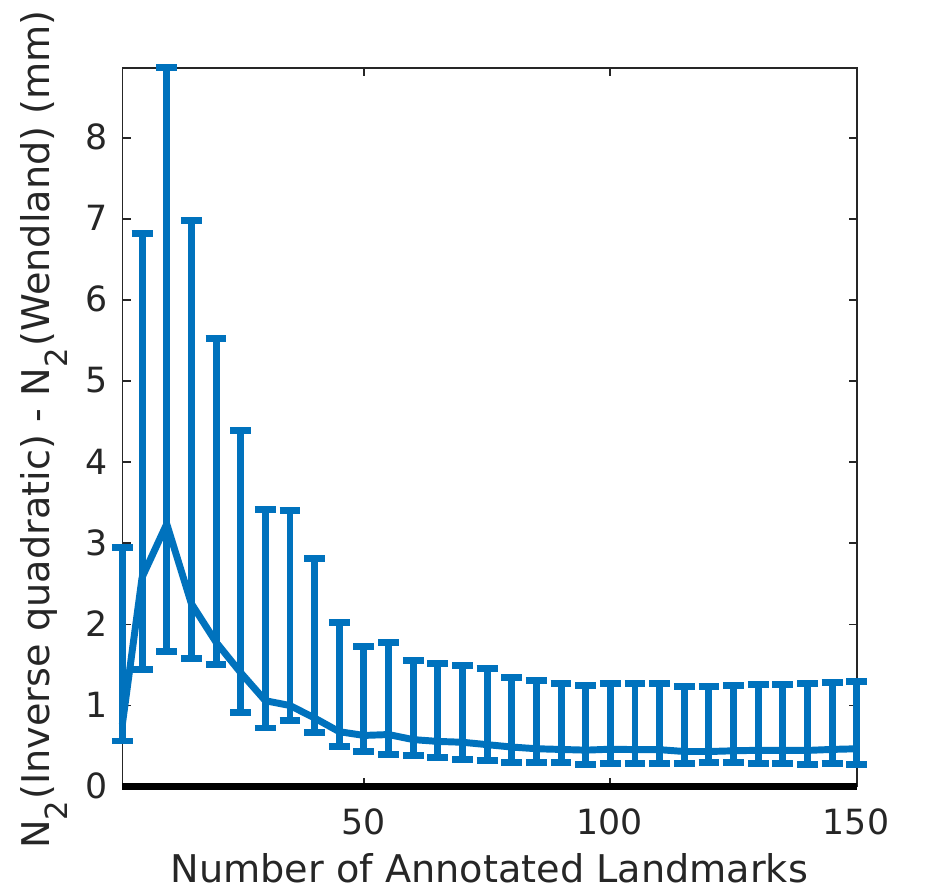}}\hfill
\caption{\textbf{Comparison between three radial basis functions.} Wendland functions yield a slightly better accuracy than the Gaussian and inverse quadratic functions.}
\label{fig:results_comparison_kernels}
\end{figure}

We measured the mean accuracy of the transformation obtained on the left-out evaluation set for different choices of basis function in our kernel~(\ref{eq:kernel_bundle}). The three radial basis functions presented in Section~\ref{sec:kernel_choice_and_parameter_estimation} were considered. The results (Fig.~\ref{fig:results_comparison_kernels}) show that the Gaussian and Wendland functions perform very similiarly and slightly better than the inverse quadratic function. When looking more precisely at the statistical distribution of the errors (Fig.~\ref{fig:results_comparison_kernels}, second and third rows), Wendland functions seem to be a slightly better choice, especially on the CIMA dataset and for a low number of annotations on the COPDgene dataset.

\subsubsection{Multi-Scale Kernel}

\begin{figure}[t]
\centering
\captionsetup{position=top}
\subfloat[CIMA dataset]
{\includegraphics[width=0.23\textwidth]{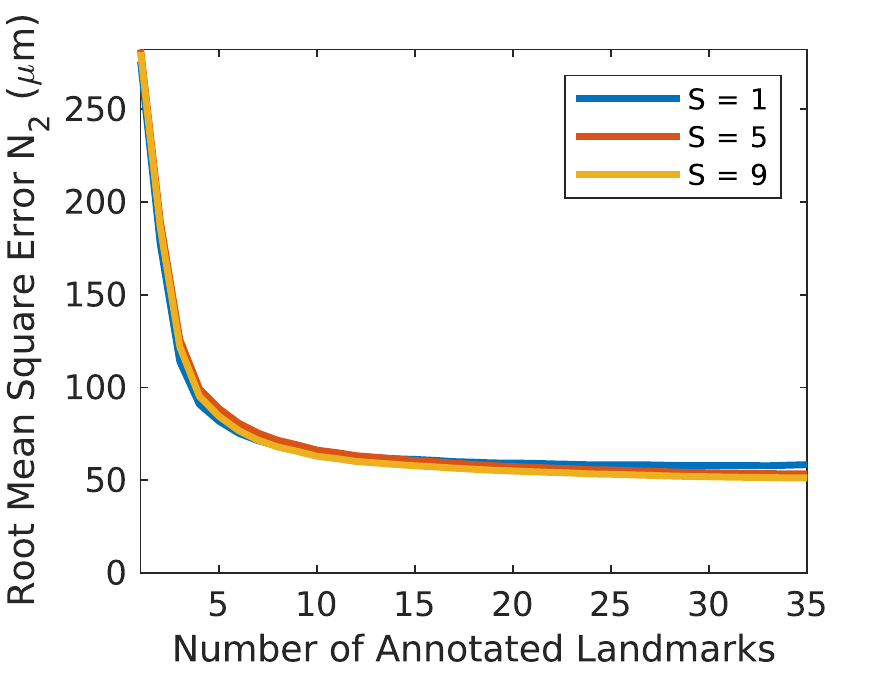}}
\subfloat[COPDgene dataset]{\includegraphics[width=0.23\textwidth]{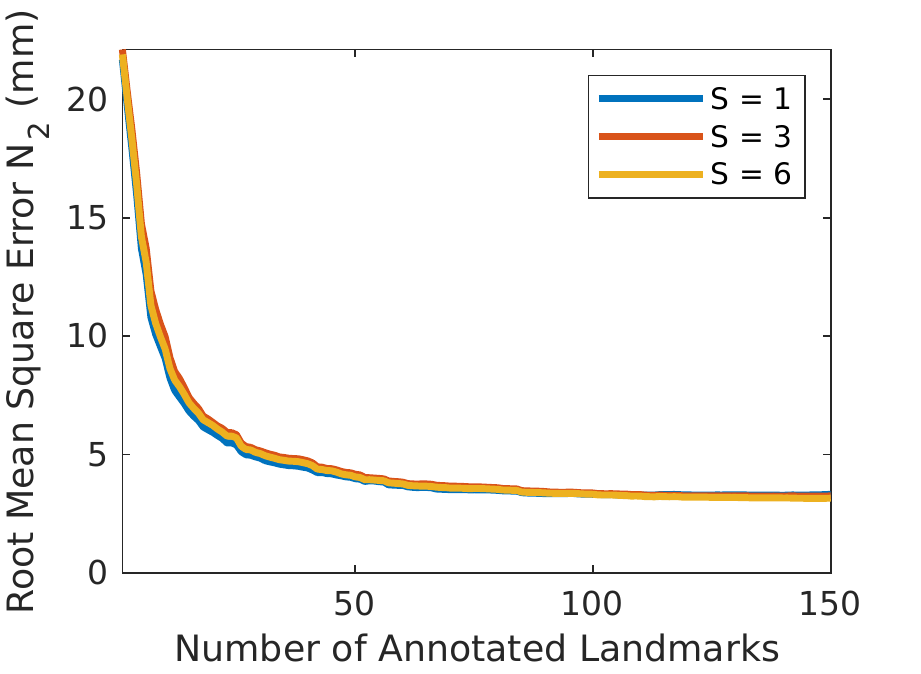}}\hfill
\subfloat{\includegraphics[width=0.23\textwidth]{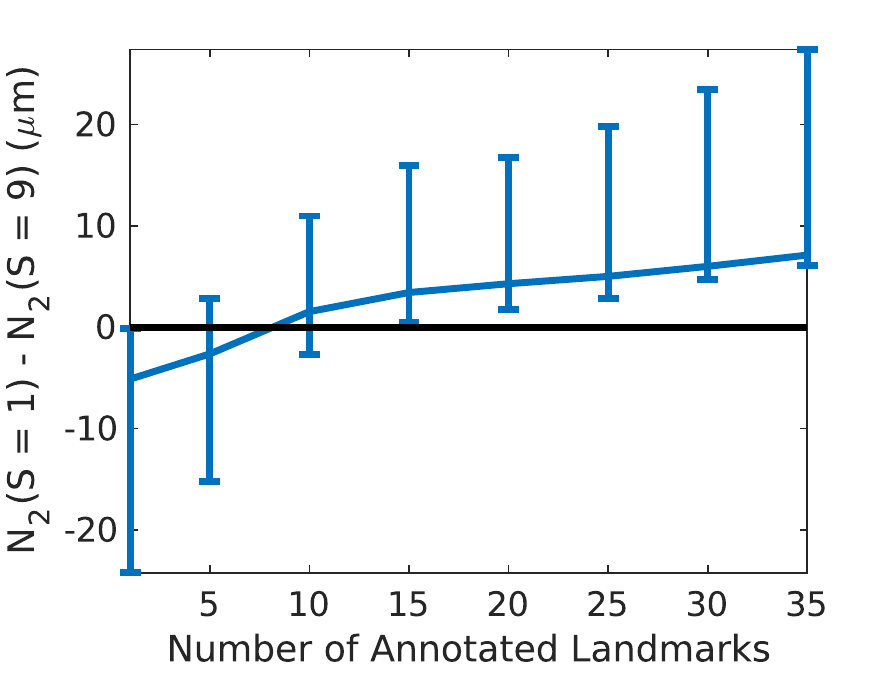}}
\subfloat{\includegraphics[width=0.23\textwidth]{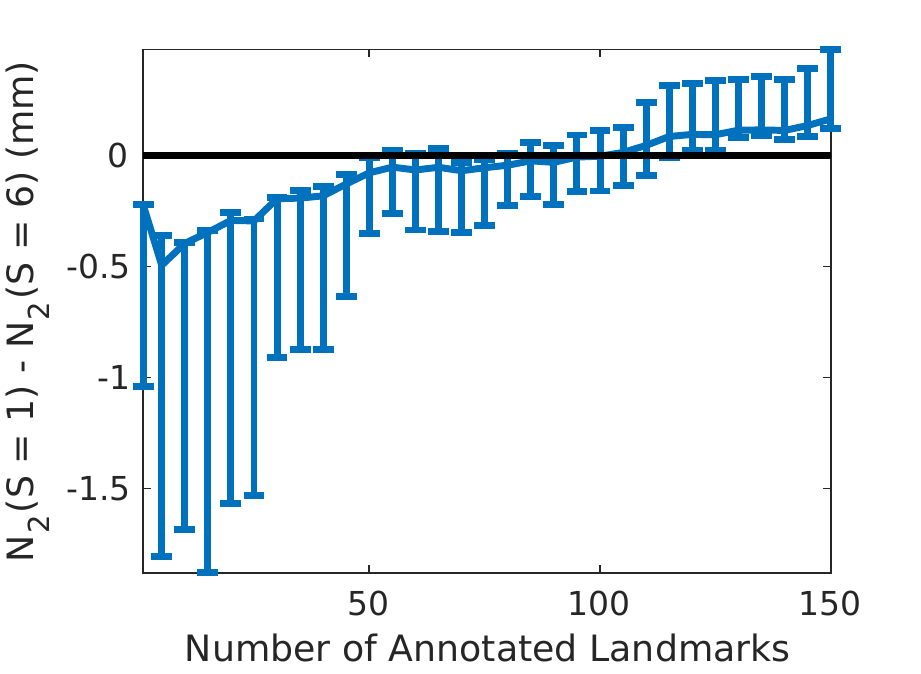}}\hfill
\subfloat{\includegraphics[width=0.23\textwidth]{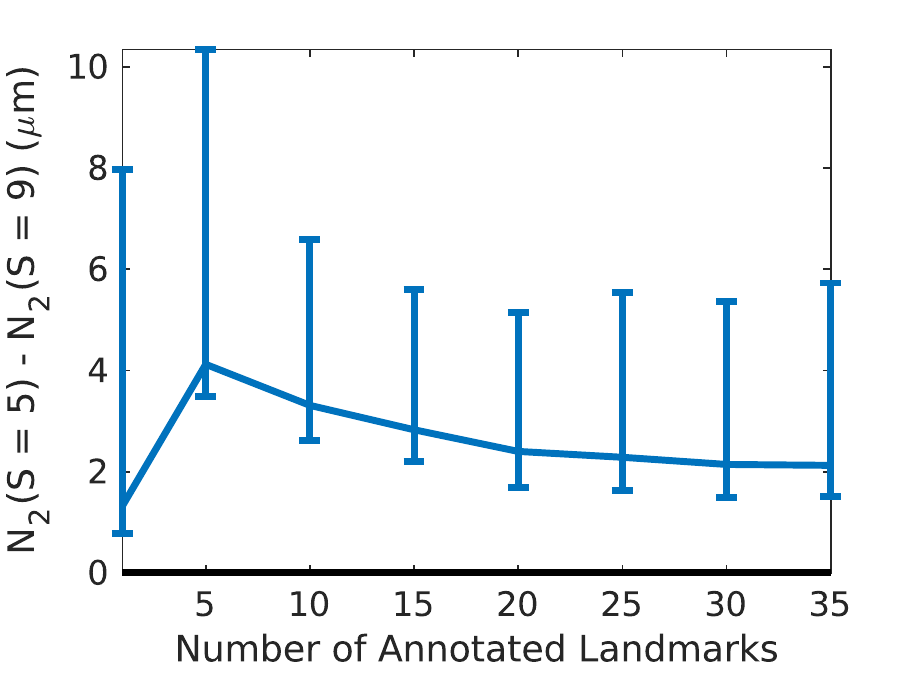}}
\subfloat{\includegraphics[width=0.23\textwidth]{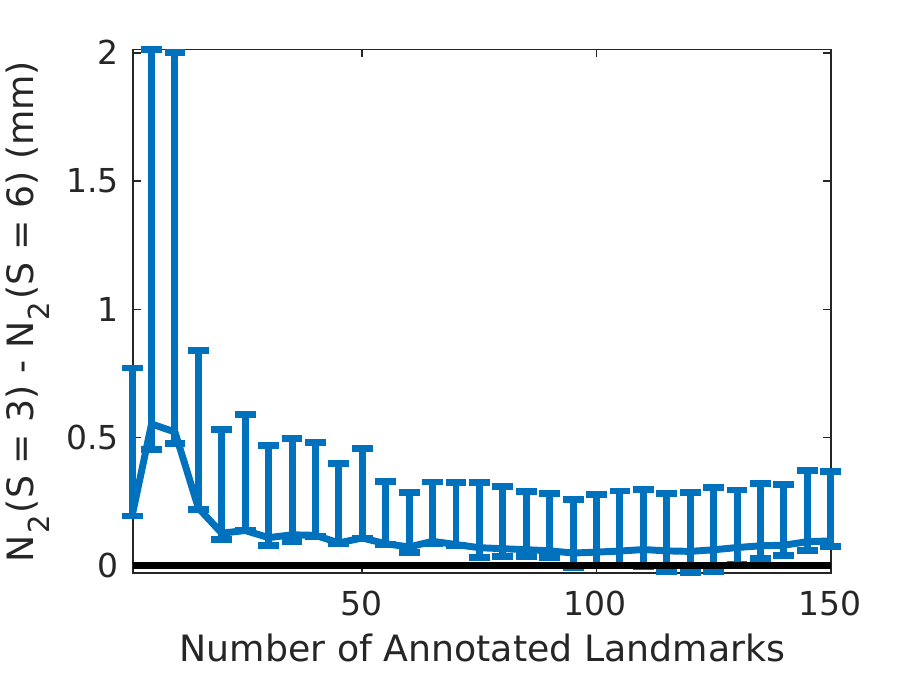}}\hfill
\caption{\textbf{Benefit of a multi-scale bundle.} We run experiments on the 2D CIMA dataset and on the 3D COPDgene dataset to study the effect of the number of scales $S$ within the multi-scale covariance function proposed in Section~\ref{sec:kernel_choice_and_parameter_estimation}. The first row shows the accuracy of the mean transformation obtained using only one scale ($S = 1$), using an intermediate value ($S = 5$ and $S = 3$, respectively), and using all scales ($S = 9$ and $S = 6$, respectively), i.e., our proposed bundle. The second and third rows plot the statistical distribution of the improvement yielded by our bundle in comparison to the single-scale kernel and the intermediate one, respectively.}
\label{fig:results_multiple_scales}
\end{figure}

Our covariance function (\ref{eq:kernel_bundle}) is defined as a multi-scale bundle of $S$ kernels acting at different scales. To motivate this design choice, we run experiments comparing our multi-scale bundle with a single-scale covariance function where only the coarsest scale is retained ($S = 1$), and with an intermediate bundle with half the number of scales (starting from the coarsest scale). The results shown in Fig.~\ref{fig:results_multiple_scales} indicate that a single coarse scale model converges more quickly but reaches a plateau of performance. Intuitively, a coarse-scaled model quickly converges to a mean transformation as defined by the landmarks, but lacks the flexibility to accurately model local deformations. As the number of landmarks increases, our multi-scale model reaches a lower error. 

\section{Conclusion and Future Work}
\label{sec:conclusion}

In the context of deformable registration, we proposed a principled probabilistic framework based on Gaussian processes for data annotation and experimental validation of algorithms. We introduced an annotation strategy to sequentially suggest the most informative locations to annotate, which yields a more accurate estimation of the true transformation than provided by a random selection or by a heuristic uniformly distributing landmarks over the image domain.
Building on our probabilistic model, we also proposed a new evaluation strategy improving over the standard landmark-based evaluation.
Future work directions include the incorporation of image-related or domain knowledge within a kernel function (e.g., using spatially varying kernels to model discontinuities), and the computationally efficient use of more complex information-theoretic criteria (e.g., mutual information~\cite{krause2008near}) to guide user interactions.

\bibliographystyle{plain}
\bibliography{LandmarkPlacement}

\end{document}